\documentclass[preprint,nonatbib]{article}
\usepackage[nonatbib]{neurips_2026}
\usepackage[numbers]{natbib}
\usepackage[utf8]{inputenc} 
\usepackage[T1]{fontenc}    
\usepackage{hyperref}       
\usepackage{url}            
\usepackage{booktabs}       
\usepackage{amsfonts}       
\usepackage{nicefrac}       
\usepackage{microtype}      
\usepackage{xcolor}         
\usepackage{graphicx}
\usepackage{algorithm}
\usepackage[noend]{algpseudocode}
\usepackage{multirow}
\usepackage{wrapfig}
\usepackage{titlesec}
\usepackage{array}
\usepackage{colortbl}
\usepackage{pifont}
\usepackage{tcolorbox}

\algrenewcommand\algorithmiccomment[1]{\hfill$\triangleright$ #1}
\titlespacing*{\paragraph}{0pt}{0.5ex plus 0.2ex minus 0.1ex}{0.7em}
\newcommand{\cmark}{\textcolor{green!50!black}{\ding{51}}}
\newcommand{\nafield}{\textcolor{gray}{--}}
\newtcolorbox{hmacebluebox}[1][]{colback=blue!4,colframe=blue!70!black,colbacktitle=blue!70!black,coltitle=white,fonttitle=\bfseries,boxrule=0.8pt,arc=2mm,left=1mm,right=1mm,top=1mm,bottom=1mm,#1}
\newtcolorbox{hmacegreenbox}[1][]{colback=green!4,colframe=green!45!black,colbacktitle=green!45!black,coltitle=white,fonttitle=\bfseries,boxrule=0.8pt,arc=2mm,left=1mm,right=1mm,top=1mm,bottom=1mm,#1}
\newtcolorbox{hmaceorangebox}[1][]{colback=orange!5,colframe=orange!75!black,colbacktitle=orange!80!black,coltitle=white,fonttitle=\bfseries,boxrule=0.8pt,arc=2mm,left=1mm,right=1mm,top=1mm,bottom=1mm,#1}

\newcommand{\algbox}[1]{%
  \Statex \hspace{\algorithmicindent}%
  \begingroup
  \setlength{\fboxsep}{3pt}%
  \fcolorbox{black!35}{white}{%
    \parbox{\dimexpr\linewidth-\algorithmicindent-2\fboxsep-2\fboxrule\relax}{%
      \raggedright\scriptsize #1}}%
  \endgroup
}

\title{HMACE: Heterogeneous Multi-Agent Collaborative Evolution for Combinatorial Optimization}

\author{%
  Yuping Yan, Jirui Han, Fei Ming, Yuanshuai Li, and Yaochu Jin \thanks{Corresponding author} \\
  School of Engineering\\
  Westlake University\\
  \texttt{yanyuping(hanjirui,mingfei,liyuanshuai,jinyaochu)@westlake.edu.cn} \\
}

\begin{document}

\maketitle

\begin{abstract}
Large Language Models have recently emerged as a promising paradigm for automated heuristic design for NP-hard combinatorial optimization problems. Despite this progress, existing LLM-based methods typically rely on monolithic workflows constrained by rigid templates, thereby restricting memory-guided exploration and triggering premature convergence to local optima. To design an autonomous and collaborative architecture, we introduce HMACE, a Heterogeneous Multi-Agent Collaborative Evolution framework that reconceptualizes heuristic search as an organizational design problem. HMACE decomposes each evolutionary generation into an autonomous, role-specialized loop with four coordinated agents: a Proposer for strategy exploration, a Generator for executable heuristic synthesis, an Evaluator for empirical assessment, and a Reflector for archive-backed memory update. By coupling behavior-aware retrieval, lightweight candidate filtering, and fitness-grounded archive updates, HMACE guides the search toward diverse and promising heuristic behaviors while avoiding redundant evaluations. Extensive evaluations on representative COPs, including TSP, Online BPP, MKP, and PFSP, show that HMACE achieves a favorable quality-efficiency trade-off compared to state-of-the-art single-agent and multi-agent baselines. In the matched LLM-driven reference comparison, HMACE achieves the lowest average gaps on TSP and Online BPP (0.464\% and 0.223\%, respectively), while requiring only 0.13M and 0.42M tokens for the two tasks, substantially fewer than the compared baselines. 
\end{abstract}

\section{Introduction}

Combinatorial Optimization Problems (COPs), such as the Traveling Salesman Problem (TSP), Vehicle Routing Problem (VRP), and Bin Packing Problem (BPP), are fundamental across operational and engineering domains \cite{alanzi2025solving, ausiello2012complexity}. However, due to their NP-hard nature \cite{papadimitriou1998combinatorial}, the feasible search space often grows exponentially with problem scale, making the acquisition of optimal solutions computationally prohibitive. Consequently, heuristic and meta-heuristic algorithms have become the prevailing approach to finding high-quality approximate solutions within acceptable timeframes \cite{rardin2001experimental}. Despite their effectiveness, traditional heuristics rely heavily on hand-crafted rules tailored by domain experts, making them costly to develop and difficult to generalize across tasks \cite{ss2022nature}.

Recently, the emergence of Large Language Models (LLMs) has revolutionized this paradigm. By leveraging their pretrained knowledge, LLMs can either directly generate promising solutions or automatically design novel heuristic policies without explicit algorithmic instructions \cite{liu2024evolution, wu2025efficient, yao2025multi}. Despite their potential, current LLM-based heuristic design methods face two fundamental structural bottlenecks. First, lacking structured exploration mechanisms such as systematic backtracking or diversification, they are highly susceptible to premature convergence and local optima \cite{liu2025experience, shigeneralizable}. Second, solving complex COPs inherently demands heterogeneous cognitive operations (e.g., planning, generation, evaluation, and refinement). Forcing a monolithic agent to unify these diverse roles inevitably triggers severe role interference \cite{li2026single}. Taken together, these limitations suggest that the bottleneck of current LLM-based solvers lies not only in model capability, but more fundamentally in the lack of appropriate organizational structure for coordinating reasoning and search.

\begin{figure*}[t]
    \centering
    \includegraphics[width=\textwidth]{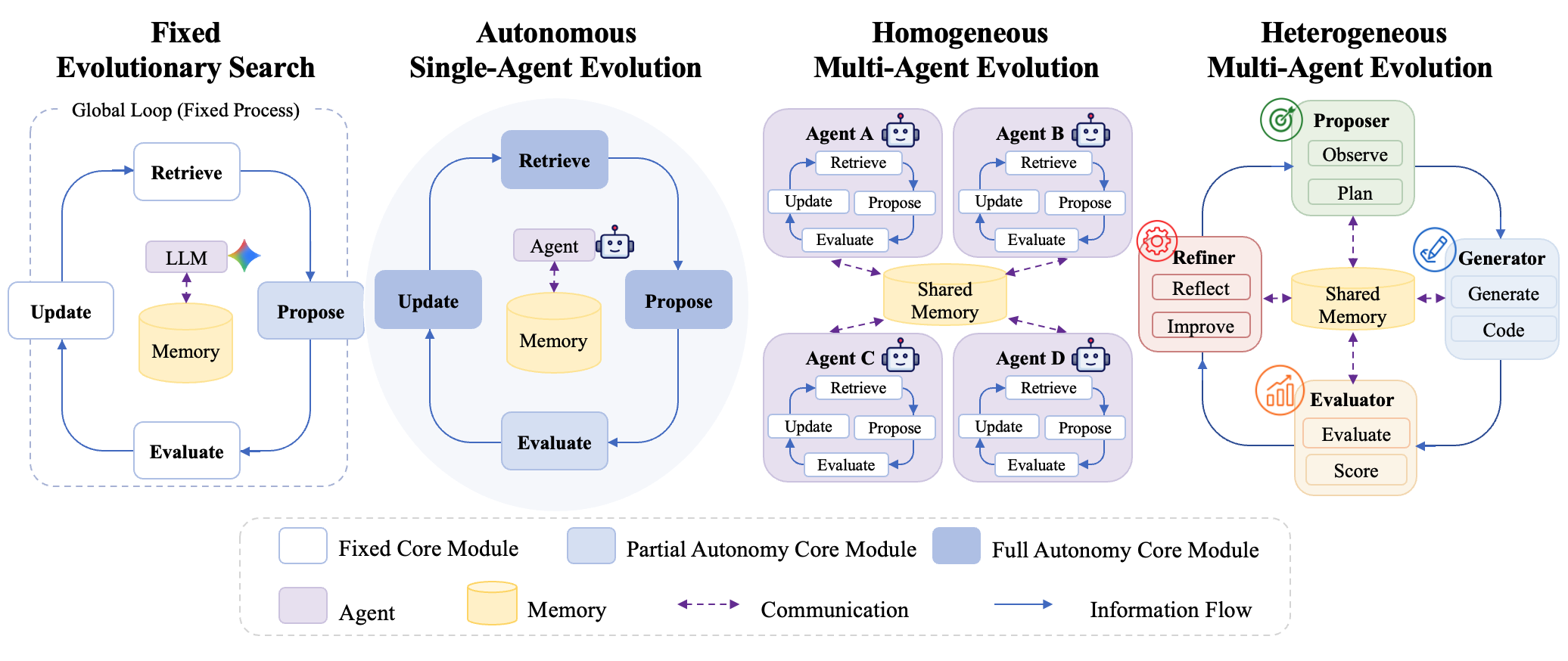}
    \caption{Architectural progression of LLM-based solvers for COPs. (1) Fixed Evolutionary Search: The LLM acts merely as a generation module within a rigid, human-crafted global loop. (2) Autonomous Single-Agent Evolution: A monolithic agent executes all cognitive operations sequentially, which inevitably leads to cognitive entanglement. (3) Homogeneous Multi-Agent Evolution: Homogeneous agents explore in parallel with a shared memory, but still lack specialized cognitive focus. (4) Heterogeneous Multi-Agent Evolution: By decoupling problem-solving into specialized, interactive roles, it robustly eliminates role interference and enables systematic, diversified exploration.}
    \label{fig:com}
\end{figure*}

Motivated by this observation, we turn to multi-agent systems as a natural solution. Instead of treating problem-solving as a monolithic generation process, multi-agent frameworks reformulate it as an organizational design problem, where multiple autonomous agents with specialized roles collaboratively explore and refine the solution space. However, prior studies have shown that naively scaling the number of agents does not guarantee performance gains; poorly designed multi-agent systems may even underperform strong single-agent baselines due to communication overhead and coordination noise \cite{liuagentbench, li2026agencybench}. As illustrated in Figure \ref{fig:com}, tackling complex COPs requires moving beyond monolithic structures and naive parallelization. \textit{Therefore, the key challenge is not merely introducing multiple agents, but designing an autonomous, role-specialized, and memory-aware multi-agent architecture that enables effective collaboration for combinatorial optimization.}

As a key step towards this new paradigm, we introduce HMACE, a novel \textbf{H}eterogeneous \textbf{M}ulti-\textbf{A}gent \textbf{C}ollaborative \textbf{E}volution framework leveraging explicit role division to automate heuristic search for COPs. In this paper, we reconceptualize the complex heuristic generation process not as a sequential coding task, but as an organizational design problem. Technically, HMACE implements a collaborative evolutionary loop to encourage both diversified exploration and reliable state management. At each iteration, specialized agents interact to propose, generate, evaluate, and archive candidate heuristics. When the search stagnates or produces low-utility candidates, the Reflector updates the archive, retrieves behaviorally diverse exemplars, and redirects subsequent proposals toward more promising or underexplored regions. By doing so, HMACE continuously refines heuristic programs while mitigating premature convergence. Ablation studies confirm that both role specialization and archive-guided reflection contribute to the observed performance gains.

Our contributions are threefold. First, we propose HMACE, a heterogeneous multi-agent evolutionary framework for combinatorial optimization, and formulate it as an organizational approach to heuristic design that goes beyond prior monolithic LLM-driven workflows. Second, we introduce a novel reflection mechanism integrated with role-specialized collaboration, enabling agents to leverage historical trajectories through behavior-aware retrieval and fitness-grounded archive updates, thereby improving the robustness and diversity of search while mitigating premature convergence. Third, extensive experiments on classical COPs demonstrate that HMACE achieves a favorable quality-efficiency trade-off compared to strong single-agent, search-controller, and multi-agent baselines. Notably, in the matched LLM-driven reference comparison, HMACE achieves the lowest average gaps on TSP and Online BPP while requiring substantially fewer tokens than the compared baselines.

\section{Related Work}

\subsection{LLM-based Automatic Heuristic Design}

Automatic heuristic design aims to reduce the dependence of COPs on manually crafted rules and domain expertise. Early automated algorithm design and hyper-heuristic methods search over predefined algorithm components, configurations, or low-level heuristic choices, whereas recent LLM-based methods expand this space to natural-language ideas and executable programs. FunSearch evolves programs with an LLM proposer and an external evaluator, demonstrating the potential of program search for mathematical discovery and online bin packing \citep{romera2024mathematical}. EoH represents heuristic ideas as natural-language ``thoughts'' paired with executable code, and evolves them within an LLM-EC loop \citep{liu2024evolution}. ReEvo further introduces reflective evolution, using LLM-generated reflections as verbal feedback to guide heuristic improvement across multiple COP settings \citep{ye2024reevo}.

Recent work has moved beyond fixed LLM-EC loops by improving the search controller itself. HSEvo and MCTS-AHD explore hierarchical or tree-structured heuristic search, while MoH explicitly studies meta-optimization of the heuristic optimizer rather than assuming a fixed evolutionary controller \citep{dat2025hsevo,zheng2025monte,shigeneralizable}. These methods suggest that the organization of the search process plays an important role in LLM-based heuristic discovery. However, they mainly improve the optimizer or search policy, rather than explicitly organizing the process as a role-specialized multi-agent system. HMACE shares the goal of automatic heuristic evolution, but emphasizes specialized agent roles, memory-aware reuse, and cost-aware filtering within a collaborative search process.

\subsection{LLMs for Combinatorial Optimization}

Beyond automatic heuristic design, LLMs have been used as modeling assistants, solver interfaces, and direct reasoning engines for combinatorial optimization. A major line of work uses LLMs to translate natural-language problem descriptions into mathematical formulations or solver-ready code. OptiMUS combines LLM-based modeling, debugging, testing, and execution with linear programming and mixed-integer linear programming solvers \citep{ahmaditeshnizi2024optimus}, while LM4OPT and LLMOPT study how language models can bridge informal specifications and formal optimization models \citep{ahmed2024lm4opt,jiang2024llmopt}. Another line of work investigates direct or assisted COP solving, covering scheduling, routing, pathfinding, and end-to-end LLM solvers \citep{abgaryan2024llms,huang2025multimodal,andreychuk2025mapf,jiang2025large,da2025large}. These works broaden the role of LLMs in optimization, but they primarily target formulation, instance-level reasoning, or solver orchestration. In contrast, HMACE does not ask the LLM to solve each instance directly; it uses LLM agents to discover reusable executable heuristics under empirical evaluation and budget constraints.

\subsection{LLM-based Multi-Agent Systems}

LLM-based multi-agent systems decompose complex tasks into interacting roles and workflows. Representative frameworks such as CAMEL, AutoGen, MetaGPT, and AgentVerse show that role-playing, tool-augmented conversation, standard operating procedures, and dynamic collaboration can improve complex reasoning and generation tasks \citep{li2023camel,wu2024autogen,hong2023metagpt,chen2023agentverse}. More closely related to optimization, DRAGON introduces decomposition and reconstruction agents for large-scale COPs, while CORAL studies autonomous multi-agent evolution for open-ended discovery \citep{chen2026dragon,qu2026coral}. However, existing multi-agent optimization frameworks do not directly address the setting of reusable heuristic discovery: DRAGON \citep{chen2026dragon} focuses on decomposing and reconstructing individual problem instances, whereas CORAL \citep{qu2026coral} targets broader open-ended discovery rather than COP-specific heuristic evolution. HMACE instead treats automatic heuristic design as a role-specialized collaborative search process, where proposal, code generation, empirical evaluation, and archive-guided reflection are coordinated under limited LLM budgets.

\section{Methodology}
\label{sec:method}

\begin{figure*}[t]
    \centering
    \includegraphics[width=\textwidth]{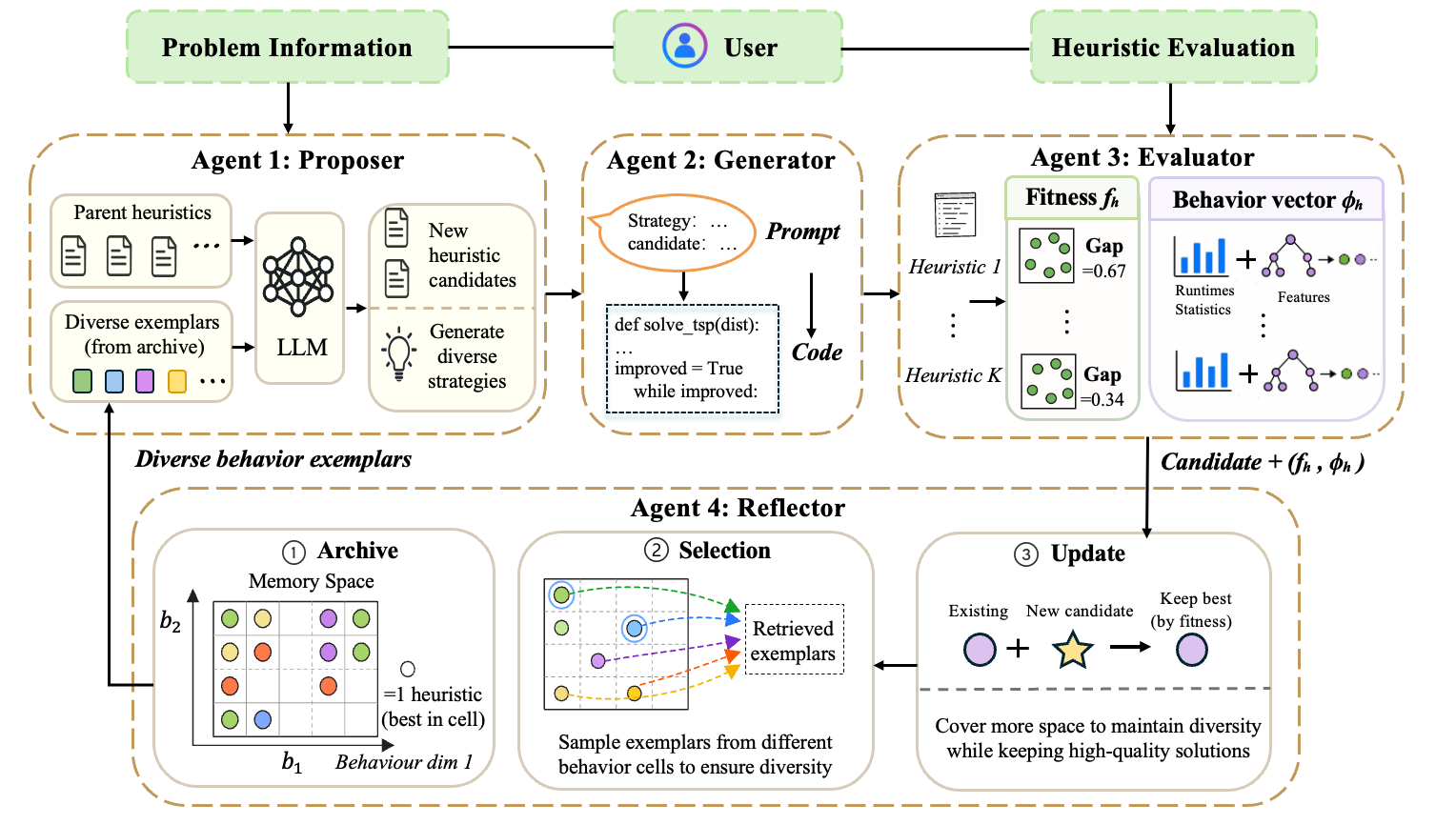}
    \caption{The overview of the HMACE framework. The proposer conditions on parent heuristics and diverse exemplars retrieved from the archive to draft strategy candidates; the generator translates each candidate into executable code; the evaluator scores candidate heuristics on the training set; and the reflector maintains an archive-backed memory, retrieves diverse exemplars from different behavioral cells, and updates each cell with the best-performing
    candidate.}
    \label{fig:hmace-method}
\end{figure*}

Figure~\ref{fig:hmace-method} presents an overview of HMACE, a heterogeneous multi-agent evolutionary framework for automatic heuristic
search on COPs. The core principle of HMACE is to decompose one search generation into four coordinated roles: a Proposer, a Generator, an Evaluator, and a Reflector, all of which communicate through an archive-based memory. In each generation, the proposer drafts strategy candidates conditioned on the current population and retrieved archive exemplars, the generator translates these strategies into executable heuristics, the evaluator measures their empirical performance on the training instances, and the reflector updates the archive to guide subsequent search. This staged design preserves diversity across behaviorally distinct niches while keeping the optimization loop grounded in measured fitness rather than verbal self-assessment. Illustrative examples of the prompts and archive records are provided in Appendix~\ref{app:prompt-design}, especially Table~\ref{tab:hmace-prompt-fields} and the prompt and memory templates collected in that subsection. In the following subsections, we present the technical details of the proposed HMACE framework.

\subsection{Problem Formulation}
\label{sec:method:formulation}

Let $\mathcal{T}$ denote a set of training
instances of a CO problem. A
\emph{heuristic} $h \in \mathbb{H}$ is an executable Python program that, given an instance state, returns a construction or repair decision. The heuristic-design objective is to identify an optimal heuristic $h^{\star} \in \mathbb{H}$, which minimizes the expected loss over $\mathcal{T}$:
\begin{equation}
\label{eq:design}
h^\star \;=\; \arg\min_{h \in \mathbb{H}} \, f(h, \mathcal{T}),
\qquad
f(h, \mathcal{T}) \;=\; \frac{1}{|\mathcal{T}|}\sum_{T \in \mathcal{T}} \ell(h, T),
\end{equation}
where $\ell(h, T)$ is the per-instance loss (excess over a lower bound or optimality gap, depending on the problem). The hypothesis class $\mathbb{H}$ is implicit, defined as the support of an LLM-driven search procedure that emits source code conditioned on the problem context.

We address Equation~\ref{eq:design} through an evolutionary paradigm. Starting from a population $P_0 \subset \mathbb{H}$, the process iteratively proposes new heuristics and selects top-performing elites evaluated under $f$. Rather than compressing an entire generation into a single, overloaded LLM call \citep{liu2024evolution, ye2024reevo, romera2024mathematical}, we decompose the generation process into the following specialized, agent-level state transitions:
\begin{equation}
\label{eq:decomp}
(P_t, \mathcal{M}_t)
\;=\;
\mathcal{R}\Bigl(
\mathcal{E}\bigl(
\mathcal{G}(\mathcal{P}(P_{t-1}, \mathcal{M}_{t-1})),\; \mathcal{T}
\bigr),\;
P_{t-1},\; \mathcal{M}_{t-1}
\Bigr),
\end{equation}
where $\mathcal{P}$ is a \emph{Proposer} drafting strategies from historical memory, $\mathcal{G}$ is a \emph{Generator} implementing these strategies as code, $\mathcal{E}$ is an \emph{Evaluator} scoring candidates on training instances, $\mathcal{R}$ is a \emph{Reflector} updating the archive to guide future iterations and $\mathcal{M}_t$ is the memory in $t$ rounds. After $G$ generations, the pipeline returns the incumbent $\hat{h}_{G} = \arg\min_{h \in P_G \cup \mathcal{M}_G} f(h, \mathcal{T})$, which serves as the empirical approximation to the ideal optimum $h^{\star}$. 
\subsection{Overall Workflow}
\label{sec:method:overview}

We summarize the \textsc{HMACE} training workflow in
the abridged Alg.~\ref{alg:hmace-workflow}; the full step-by-step
pseudocode is deferred to Appendix~\ref{app:full-algorithm}
(Alg.~\ref{alg:hmace-workflow-full}). Algorithm~\ref{alg:hmace-workflow} writes the
workflow as an outer evolutionary loop wrapped around a generation
routine \textsc{GenStep}$(P,\mathcal{M})$. Starting from a seed-induced
population $P_{0}$ and an empty archive-backed memory $\mathcal{M}_{0}$,
HMACE iteratively alternates between retrieval, proposal-generation,
evaluation, and reflection.

Within each generation, the reflector first performs retrieval through
$\mathcal{R}_{\mathrm{sel}}$, which samples $r$ diverse exemplars from the
current population and occupied archive cells to form the context
exposed to the proposer. Conditioned on this context, the
\textbf{Proposer} drafts $k$ strategy candidates, and the
\textbf{Generator} translates them into executable heuristics. Before
expensive evaluation, a lightweight deterministic filter removes
malformed or near-duplicate programs so that the search budget is spent
on feasible and behaviorally non-redundant candidates. The \textbf{Evaluator} then assigns each surviving heuristic two
complementary signals: a scalar fitness score $f_h = f(h,\mathcal{T})$
that measures solution quality on the training set, and a behavior
vector $\phi_h = \phi(h,\mathcal{T})$ that characterizes how the
heuristic behaves during execution. These evaluated tuples
$(h,\phi_h,f_h)$ are passed to the reflector update step
$\mathcal{R}_{\mathrm{upd}}$, which refreshes the population by fitness,
updates archive incumbents cell by cell, and returns the next search
state $(P_t,\mathcal{M}_t)$. The outer loop repeats this generation
routine until the budget of $G$ generations is exhausted or the best
fitness fails to improve by at least $\delta_{\min}=10^{-4}$ for
$\rho{=}3$ consecutive generations. In this sense,
reflection in HMACE is not merely verbal critique; it is an operational
memory mechanism that jointly controls exploitation through fitness and
exploration through behavior-space coverage.

\begin{algorithm}[t]
\caption{Abridged \textsc{HMACE} training workflow.}
\label{alg:hmace-workflow}
\footnotesize
\begin{algorithmic}[1]
  \Statex \textbf{Input:} number of generations $G$, population capacity $n$, proposal batch size $k$, retrieval size $r$, archive capacity $C$, patience $\rho$, minimum improvement threshold $\delta_{\min}$, training instances $\mathcal{T}$, seed prompt $\Pi_{0}$.
  \Statex \textbf{Output:} best heuristic $h^{\star}$, final population $P_{G}$, shared memory $\mathcal{M}_{G}$.
  \Function{\textsc{GenStep}}{$P,\mathcal{M}$}
    \State $X \gets \mathcal{R}_{\mathrm{sel}}(P,\mathcal{M},r)$ \label{line:alg-retrieval-start}
    \State $S = \{s_i\}_{i=1}^{k} \gets \mathcal{P}(P,X)$ \label{line:alg-proposal-start}
    \State $H = \{h_i\}_{i=1}^{k} \gets \mathcal{G}(S)$ \label{line:alg-generation-start}
    \State $\widetilde{H} \gets \textsc{Filter}(H,\mathcal{M})$ \label{line:alg-filter-end}
    \For{$h \in \widetilde{H}$}
      \State $f_h \gets f(h,\mathcal{T}),\qquad \phi_h \gets \phi(h,\mathcal{T})$ \label{line:alg-eval-loop}
    \EndFor
    \State $Q \gets \{(h,\phi_h,f_h): h \in \widetilde{H}\}$ \label{line:alg-eval-end}
    \State $(P,\mathcal{M}) \gets \mathcal{R}_{\mathrm{upd}}(P,\mathcal{M},Q)$ \label{line:alg-update-start}
    \State \Return $(P,\mathcal{M})$ \label{line:alg-genstep-end}
  \EndFunction
  \State $P_{0} \gets \textsc{LLM}(\Pi_{0})$, $\mathcal{M}_{0} \gets \emptyset$ \label{line:alg-init}
  \For{$t = 1,\ldots,G$} \label{line:alg-outer-loop-start}
    \State $(P_{t},\mathcal{M}_{t}) \gets \textsc{GenStep}(P_{t-1},\mathcal{M}_{t-1})$ \label{line:alg-genstep-call}
    \If{$\textsc{NoImprovement}(P_{t},\rho,\delta_{\min})$}
      \State \textbf{break}
    \EndIf
  \EndFor
  \State $h^{\star} \gets \arg\min_{(h,\phi,f)\in P_{G}\cup \mathcal{M}_{G}} f$ \label{line:alg-select-best}
  \State \Return $(h^{\star},P_{G},\mathcal{M}_{G})$ \label{line:alg-return}
\end{algorithmic}
\smallskip
\end{algorithm}

\subsection{Detailed Implementation}
\label{sec:method:implementation}

As key implementation stages of HMACE, we further elaborate on the
generation routine, especially the retrieval, proposal-generation,
evaluation, and memory-update blocks in
lines~\ref{line:alg-retrieval-start}--\ref{line:alg-genstep-end} of
\textsc{GenStep}$(P,\mathcal{M})$. The main components are as follows.

\paragraph{Initialization and shared memory.} The initialization block
(lines~\ref{line:alg-init}--\ref{line:alg-genstep-call}) initializes HMACE with a seed-induced population
$P_{0} \gets \textsc{LLM}(\Pi_{0})$ and an empty archive-backed memory
$\mathcal{M}_{0}$. Rather than storing a flat interaction history,
$\mathcal{M}$ stores evaluated tuples $(h,\phi_h,f_h)$, where $h$ is an
executable heuristic, $f_h = f(h,\mathcal{T})$ is its fitness, and
$\phi_h = \phi(h,\mathcal{T}) \in \mathbb{R}^{d}$ is a behavior
descriptor derived from execution. The shared memory therefore simultaneously preserves high-quality heuristics and exposes the behavioral structure needed for diversity-aware retrieval.

\paragraph{Retrieval and strategy proposal.} At the beginning of each
generation, the reflector executes
$\mathcal{R}_{\mathrm{sel}}(P,\mathcal{M},r)$ to construct the exemplar
set $X$. In the exemplar-retrieval and proposal block (lines~\ref{line:alg-retrieval-start}--\ref{line:alg-proposal-start}), this step
builds a candidate pool from the current population and archive elites,
then selects $r$ exemplars from behaviorally distinct occupied cells.
The proposer $\mathcal{P}$ receives both the parent heuristics in $P$
and the retrieved set $X$, and drafts $k$ candidate strategies in natural language $S = \{s_i\}_{i=1}^{k}$. Because retrieval is conditioned on
different behavioral niches, proposal is encouraged to combine,
contrast, or extend distinct search patterns rather than refine a single dominant heuristic family.

\paragraph{Code generation and pre-filtering.} The generator
$\mathcal{G}$ maps each candidate strategy $s_i$ to an executable
heuristic $h_i$ that respects the problem description and task IO
contract. HMACE then applies the lightweight routine
$\textsc{Filter}(H,\mathcal{M})$ before expensive evaluation. This
deterministic guardrail removes malformed, infeasible, or near-duplicate
programs using contract checks, basic syntax or signature validation,
and duplicate detection against the current batch.
Consistent with the code-generation and pre-filtering block (lines~\ref{line:alg-generation-start}--\ref{line:alg-filter-end}), this filter is not a fifth reasoning agent but rather an implementation-level screen designed to conserve the search budget by rejecting invalid or redundant candidates.

\paragraph{Evaluation and behavior characterization.} For each
surviving heuristic $h \in \widetilde{H}$, the evaluator
$\mathcal{E}$ executes $h$ on the training set $\mathcal{T}$ and
returns two complementary signals: the scalar fitness score
$f_h = f(h,\mathcal{T})$, which measures solution quality, and the
behavior vector $\phi_h = \phi(h,\mathcal{T})$, which summarizes how
the heuristic acts during execution. The resulting set
$Q = \{(h,\phi_h,f_h)\}$ grounds the search in measurable performance
rather than self-reported quality and provides the information needed
for both elite selection and archive placement.

\paragraph{Population and archive update.} The reflector then invokes
$\mathcal{R}_{\mathrm{upd}}(P,\mathcal{M},Q)$ to refresh the search
state. In the population-update block (lines~\ref{line:alg-update-start}--\ref{line:alg-genstep-end}), it updates the
population by fitness as $P \gets \mathrm{Top}_{n}(P \cup \{(h,f_h):(h,\phi_h,f_h)\in Q\})$, inserts each evaluated candidate into its behavior cell, and keeps the best incumbent in that niche while maintaining archive capacity $C$ and persisting $\mathcal{M}$ for resume safety. This rule couples exploitation and exploration: fitness controls which heuristics survive in the population, whereas behavior descriptors control archive occupancy and coverage over the search space, following the quality-diversity intuition of archive-based search
methods \citep{mouret2015illuminating, vassiliades2017cvt}. The outer
loop repeats until the generation budget $G$ is exhausted or
$\textsc{NoImprovement}(P_t,\rho)$ triggers early stopping.

\paragraph{Inter-agent coordination and computational profile.} The typed interfaces across the staged pipeline (lines~\ref{line:alg-retrieval-start}--\ref{line:alg-return}) are intentionally narrow, where the proposer emits strategies, the generator emits executable heuristics, the evaluator emits $(f_h,\phi_h)$ pairs, and the reflector emits retrieved exemplars together with the updated $(P,\mathcal{M})$. Relative to free-form conversational multi-agent systems, these typed handoffs make failures easier to attribute and allow each role to be replaced or ablated independently. In terms of
cost, one generation uses one proposer call and up to $k$ generator calls, followed by deterministic filtering, evaluation, and archive maintenance.

\section{Experiments}

\paragraph{Tasks and Datasets.} \textbf{TSP.} We consider Hamiltonian-tour construction on $77$ EUC\_2D TSPLIB instances ranging from $50$ to $20$k cities, and use the optimal tour lengths or lower bounds reported in TSPLIB as oracle references \citep{reinelt1991tsplib}. This remains the canonical routing benchmark in both classical and neural COP research \citep{applegate2006traveling,helsgaun2000lkh,vinyals2015pointer,
bello2017neural,kool2019attention,kwon2020pomo,luo2023lehd}. \textbf{BPP.} We consider online bin packing on the Weibull-5k distribution used by FunSearch and EoH, and use the corresponding FunSearch L2 lower bounds as oracle references \citep{romera2024mathematical,liu2024evolution}. This task has long served as a benchmark for approximation algorithms, metaheuristics, and recent LLM-driven heuristic search \citep{johnson1973bin,coffman1996approximation,falkenauer1996hybrid}. \textbf{MKP.} We consider profit maximization under multiple knapsack
capacity constraints and generate $10$ synthetic instances following our
solver-based setup: capacities are sampled from $\mathcal{U}(100,500)$, the number of knapsacks varies from $10$ to $100$, and item values and weights are sampled from $\mathcal{U}(1,100)$. Oracle values are taken from the best feasible values or upper bounds returned by the exact solver \citep{martello1990knapsack,chu1998genetic,perron2024ortools}. \textbf{PFSP.} We consider permutation flow-shop scheduling on standard
Taillard benchmark instances spanning multiple job-machine scales, and use the best-known makespan values in the benchmark literature as oracle references \citep{taillard1993benchmarks}. Classical constructive baselines include NEH and Gupta \citep{nawaz1983neh,gupta1971functional}.

\paragraph{Baseline.}
We group the baselines into four categories. First, we include classical problem-specific heuristics and OR references, such as Concorde, OR-Tools, and nearest neighbor for TSP, Best Fit and First Fit for online BPP, and NEH and Gupta for PFSP \citep{applegate2006traveling,perron2024ortools,johnson1973bin,
coffman1996approximation,nawaz1983neh,gupta1971functional}. Second, we compare HMACE with representative single-agent LLM heuristic-discovery methods, namely FunSearch and EoH \citep{romera2024mathematical,liu2024evolution}. Third, we include stronger search-controller baselines that improve the optimizer itself, including ReEvo, HSEvo, MCTS-AHD, and MoH \citep{ye2024reevo,dat2025hsevo,zheng2025monte,shigeneralizable}. Finally, we report recent multi-agent references, namely CORAL and, in the TSP/BPP reference comparison, DRAGON \citep{qu2026coral,chen2026dragon}. This baseline suite allows us to compare HMACE against fixed hand-designed heuristics, representative single-agent LLM-EC methods, optimizer-level search improvements, and recent multi-agent alternatives, with additional PFSP/MKP results reported in Appendix~\ref{app:pfsp-mkp} (Table~\ref{tab:pfsp-mkp-results}) and complementary diagnostic analyses reported in Appendix~\ref{app:convergence-anytime} (Figures~\ref{fig:appendix-convergence} and \ref{fig:appendix-anytime}), Appendix~\ref{app:behavior-coverage} (Figures~\ref{fig:appendix-behavior-space} and \ref{fig:appendix-archive-coverage}), and Appendix~\ref{app:cost-frontier} (Figures~\ref{fig:appendix-ablation-visual}--\ref{fig:appendix-token-efficiency}).

\paragraph{Implementation Details.} 
All methods use a population size of 10 and a generation budget of 30. We adopt a patience-3 plateau-based early-stopping rule, i.e., optimization terminates when the best fitness fails to improve by at least $\delta_{\min}=10^{-4}$ for three consecutive generations. Each run uses $n_{\mathrm{proc}}=12$ evaluator workers in parallel. For HMACE, we use a CVT-MAP-Elites archive with $n_{\mathrm{centroids}}=25$ and a behavior descriptor of dimension $d_b=11$, comprising five runtime statistics and six static program-structure features. In each generation, the reflector retrieves $r=2$ archive exemplars for the proposer, the generator produces $k=4$ candidate children, and the lightweight screening stage retains the top 50\% of candidates before evaluation. For each problem-method pair, we run three random seeds, $\{0,1,2\}$, and report the mean $\pm$ standard deviation. We additionally log per-candidate token usage, wall-clock time, behavior vectors, archive occupancy, and screening decisions in a per-run Excel trace for downstream analysis.

\paragraph{Metrics.} We evaluate performance using four metrics: (1) \emph{relative suboptimality}, defined as $\Delta = \frac{\left| f - f^{\star} \right|}{\left| f^{\star} \right|} \times 100\%$, where $f$ and $f^{\star}$ denote the obtained objective value and the optimal objective value, respectively; (2) \emph{wall-clock runtime} (seconds); (3) \emph{input/output token usage}; and (4) \emph{the number of API
queries}. All experiments use a unified configuration with a 1-hour time budget $\tau_{\mathrm{lim}} = 3600$ and the patience-3 early-stopping rule described above.

\subsection{Empirical Results}

Table~\ref{tab:tsp-results} reports TSP. All three GPT-5.4-evaluated methods match the optimum at 20 and 50 cities. From 100 cities onward,
\textsc{HMACE}$^{*}$ becomes the best row, attaining the lowest gaps at 100, 200, 500, and 1000 cities and the best average gap overall ($0.387\%$), compared with $0.431\%$ for \textsc{CORAL}$^{*}$ and $0.467\%$ for EoH$^{*}$. This pattern indicates that the advantage of \textsc{HMACE} becomes clearer as the search space grows.

\begin{table}[H]
\caption{Results of constructive and improvement heuristics on TSP. Methods marked with $^{*}$ are evaluated by the authors using GPT-5.4.}
\label{tab:tsp-results}
\centering
\scriptsize
\setlength{\tabcolsep}{3pt}
\renewcommand{\arraystretch}{1.08}
\resizebox{\textwidth}{!}{%
\begin{tabular}{c|cc|cc|cc|cc|cc|cc|c}
\toprule
\multicolumn{1}{c|}{} & \multicolumn{8}{c|}{Train} & \multicolumn{4}{c|}{Generalization} & \multicolumn{1}{c}{Average Gap$\downarrow$} \\
\multicolumn{1}{c|}{Methods} & \multicolumn{2}{c|}{20} & \multicolumn{2}{c|}{50} & \multicolumn{2}{c|}{100} & \multicolumn{2}{c|}{200} & \multicolumn{2}{c|}{500} & \multicolumn{2}{c|}{1000} & \\
 & Obj.$\downarrow$ & Gap$\downarrow$ & Obj.$\downarrow$ & Gap$\downarrow$ & Obj.$\downarrow$ & Gap$\downarrow$ & Obj.$\downarrow$ & Gap$\downarrow$ & Obj.$\downarrow$ & Gap$\downarrow$ & Obj.$\downarrow$ & Gap$\downarrow$ & \\
\midrule
Concorde & 3.840 & - & 5.715 & - & 7.766 & - & 10.679 & - & 16.519 & - & 23.104 & - & - \\
OR-Tools & 3.840 & 0.000\% & 5.715 & 0.001\% & 7.772 & 0.089\% & 10.944 & 2.478\% & 17.259 & 4.479\% & 24.262 & 5.011\% & 2.010\% \\
Nearest Neighbor & 4.602 & 19.806\% & 7.055 & 23.406\% & 9.636 & 24.072\% & 13.374 & 25.228\% & 20.691 & 25.252\% & 28.990 & 25.474\% & 23.873\% \\
Funsearch & 4.261 & 11.000\% & 6.523 & 14.162\% & 9.018 & 16.109\% & 12.615 & 18.143\% & 19.531 & 18.242\% & 27.571 & 19.332\% & 16.165\% \\
EoH & 4.204 & 9.408\% & 6.402 & 12.007\% & 8.774 & 12.974\% & 12.233 & 14.548\% & 19.029 & 15.196\% & 26.890 & 16.390\% & 13.420\% \\
ReEvo & 4.197 & 9.250\% & 6.399 & 11.966\% & 8.786 & 13.133\% & 12.217 & 14.403\% & 19.035 & 15.232\% & 26.818 & 16.076\% & 13.343\% \\
HSEvo & 4.108 & 6.897\% & 6.280 & 9.881\% & 8.705 & 12.102\% & 12.208 & 14.320\% & 19.550 & 18.349\% & 27.431 & 18.727\% & 13.379\% \\
MCTS-AHD & 4.107 & 6.882\% & 6.332 & 10.807\% & 8.735 & 12.499\% & 12.165 & 13.921\% & 19.036 & 15.240\% & 26.814 & 16.060\% & 12.568\% \\
MoH & 4.104 & 6.837\% & 6.280 & 9.893\% & 8.654 & 11.444\% & 12.100 & 13.307\% & 18.869 & 14.224\% & 26.581 & 15.049\% & 11.792\% \\
\midrule
EoH$^{*}$ & \textbf{3.840} & \textbf{0.000\%} & \textbf{5.715} & \textbf{0.000\%} & 7.768 & 0.028\% & 10.711 & 0.301\% & 16.677 & 0.956\% & 23.455 & 1.515\% & 0.467\% \\
\textsc{CORAL}$^{*}$ & \textbf{3.840} & \textbf{0.000\%} & \textbf{5.715} & \textbf{0.000\%} & 7.769 & 0.039\% & 10.705 & 0.243\% & 16.665 & 0.882\% & 23.432 & 1.420\% & 0.431\% \\
\textsc{HMACE}$^{*}$ & \textbf{3.840} & \textbf{0.000\%} & \textbf{5.715} & \textbf{0.000\%} & \textbf{7.767} & \textbf{0.014\%} & \textbf{10.698} & \textbf{0.177\%} & \textbf{16.653} & \textbf{0.809\%} & \textbf{23.409} & \textbf{1.319\%} & \textbf{0.387\%} \\
\bottomrule
\end{tabular}}
\end{table}

Table~\ref{tab:online-bpp} reports Online BPP. Under this problem, \textsc{HMACE}$^{*}$ remains the best starred method on the average row ($0.441\%$) and outperforms MoH in 12 of the 15 settings. EoH$^{*}$ is generally slightly better than the original EoH row, while \textsc{CORAL}$^{*}$ stays between HSEvo and MoH. The overall ordering therefore remains stable, with \textsc{HMACE}$^{*}$ giving the best overall BPP performance among the three GPT-5.4 rows. Taken together, these results support our central claim: the main benefit of \textsc{HMACE} is not only stronger search quality, but also more
efficient use of evaluation budget.

Table~\ref{tab:tsp-bpp-reference} summarizes the comparative performance of HMACE against other baselines on the TSP and Online BPP tasks. As shown, HMACE achieves the best overall solution quality, yielding the lowest average objective gaps (0.464\% for TSP and 0.223\% for Online BPP) and outperforming all baseline methods by a significant margin. Furthermore, HMACE demonstrates exceptional token efficiency, requiring only 0.13M and 0.42M tokens for the two tasks, respectively. While CORAL exhibits a shorter execution time, this speed comes at the severe cost of inflated token consumption (e.g., 5.48M tokens for BPP, over 13$\times$ that of HMACE) and degraded solution quality. Overall, HMACE achieves the optimal balance of superior heuristic performance and minimal API token cost.

\begin{table}[H]
\caption{Results on Online BPP. Methods marked with $^{*}$ are evaluated by GPT-5.4.}
\label{tab:online-bpp}
\centering
\scriptsize
\setlength{\tabcolsep}{4pt}
\renewcommand{\arraystretch}{1.08}
\resizebox{\textwidth}{!}{%
\begin{tabular}{cc|cccccccc|ccc}
\toprule
Bin Capacity & Item Size & Best Fit & First Fit & FunSearch & EoH & ReEvo & HSEvo & MCTS-AHD & MoH & EoH$^*$ & CORAL$^*$ & HMACE$^*$ \\
\midrule
\multirow{3}{*}{100} & 1k  & 4.621\% & 5.038\% & 3.165\% & 3.294\% & 3.475\% & 3.748\% & \textbf{2.543\%} & 2.553\% & 3.214\% & 3.091\% & 2.595\% \\
 & 5k  & 4.149\% & 4.488\% & 2.165\% & 0.827\% & 2.022\% & 1.088\% & 1.769\% & 0.600\% & 0.801\% & 0.820\% & \textbf{0.585\%} \\
 & 10k & 4.030\% & 4.308\% & 2.008\% & 0.436\% & 1.821\% & 0.734\% & 1.647\% & 0.414\% & 0.421\% & 0.558\% & \textbf{0.404\%} \\
\midrule
\multirow{3}{*}{200} & 1k  & 1.825\% & 2.025\% & 0.938\% & 1.645\% & 1.825\% & 1.825\% & 1.238\% & 0.848\% & 1.604\% & 1.288\% & \textbf{0.819\%} \\
 & 5k  & 1.555\% & 1.665\% & 0.543\% & 0.366\% & 1.549\% & 1.555\% & 1.062\% & 0.262\% & 0.379\% & 0.844\% & \textbf{0.223\%} \\
 & 10k & 1.489\% & 1.578\% & 0.459\% & 0.188\% & 1.489\% & 1.489\% & 1.036\% & 0.141\% & 0.176\% & 0.748\% & \textbf{0.101\%} \\
\midrule
\multirow{3}{*}{300} & 1k  & 1.131\% & 1.265\% & 0.654\% & 1.086\% & 1.131\% & 1.131\% & 0.922\% & \textbf{0.581\%} & 1.042\% & 0.829\% & 0.600\% \\
 & 5k  & 0.919\% & 0.984\% & 0.352\% & 0.254\% & 0.919\% & 0.919\% & 0.785\% & 0.161\% & 0.243\% & 0.502\% & \textbf{0.138\%} \\
 & 10k & 0.882\% & 0.924\% & 0.316\% & 0.115\% & 0.882\% & 0.882\% & 0.765\% & 0.079\% & 0.118\% & 0.440\% & \textbf{0.055\%} \\
\midrule
\multirow{3}{*}{400} & 1k  & 0.815\% & 0.835\% & 0.519\% & 0.815\% & 0.815\% & 0.815\% & 0.755\% & 0.498\% & 0.792\% & 0.641\% & \textbf{0.488\%} \\
 & 5k  & 0.624\% & 0.672\% & 0.275\% & 0.191\% & 0.621\% & 0.624\% & 0.608\% & 0.104\% & 0.183\% & 0.338\% & \textbf{0.088\%} \\
 & 10k & 0.603\% & 0.639\% & 0.243\% & 0.098\% & 0.595\% & 0.603\% & 0.579\% & 0.054\% & 0.094\% & 0.301\% & \textbf{0.038\%} \\
\midrule
\multirow{3}{*}{500} & 1k  & 0.546\% & 0.522\% & \textbf{0.324\%} & 0.695\% & 0.546\% & 0.546\% & 0.496\% & 0.373\% & 0.708\% & 0.451\% & 0.379\% \\
 & 5k  & 0.472\% & 0.507\% & 0.214\% & 0.119\% & 0.472\% & 0.472\% & 0.447\% & 0.090\% & 0.111\% & 0.262\% & \textbf{0.079\%} \\
 & 10k & 0.448\% & 0.487\% & 0.196\% & 0.075\% & 0.445\% & 0.448\% & 0.430\% & 0.032\% & 0.071\% & 0.219\% & \textbf{0.020\%} \\
\midrule
\multicolumn{2}{c|}{Average} & 1.607\% & 1.729\% & 0.825\% & 0.680\% & 1.240\% & 1.125\% & 1.006\% & 0.453\% & 0.664\% & 0.755\% & \textbf{0.441\%} \\
\bottomrule
\end{tabular}}
\end{table}

\begin{table}[H]
\caption{Comparisons of TSP and Online BPP. Methods marked with $^{*}$ are evaluated using GPT-5.4. For TSP, Obj. (\%) reports the average optimality gap over the 50-, 100-, 200-, 500-, and 1000-city settings in Table~\ref{tab:tsp-results}; for Online BPP, Obj. (\%) reports the average over the Weibull-5k rows in Table~\ref{tab:online-bpp}. Gap (\%) is computed relative to the best GPT-5.4 row within each matched task block. Time (s) and Tokens report the available runtime and token logs.}
\label{tab:tsp-bpp-reference}
\centering
\scriptsize
\setlength{\tabcolsep}{4pt}
\renewcommand{\arraystretch}{1.08}
\resizebox{\textwidth}{!}{%
\begin{tabular}{c|cccc|cccc}
\toprule
\multicolumn{1}{c|}{Methods} & \multicolumn{4}{c|}{TSP ($\downarrow$)} & \multicolumn{4}{c}{Online BPP ($\downarrow$)} \\
Methods & Obj. (\%) & Gap (\%) & Time (s) & Tokens & Obj. (\%) & Gap (\%) & Time (s) & Tokens \\
\midrule
EoH$^*$ & 0.560 & 20.7 & 3543 & 0.24M & 0.343 & 54.3 & 12462 & 1.29M \\
CORAL$^*$ & 0.517 & 11.4 & \textbf{1111} & 0.94M & 0.453 & 68.5 & \textbf{1191} & 5.48M \\
DRAGON$^*$ & 0.572 & 23.3 & 2493 & 0.26M & 0.337 & 51.1 & 12108 & 1.24M \\
\midrule
HMACE$^*$ & \textbf{0.464} & \textbf{0.00} & 1186 & \textbf{0.13M} & \textbf{0.223} & \textbf{0.00} & 4024 & \textbf{0.42M} \\
\bottomrule
\end{tabular}}
\end{table}

\begin{wrapfigure}[14]{r}{0.53\textwidth}
\vspace{-1.2ex}
\centering
\includegraphics[width=\linewidth]{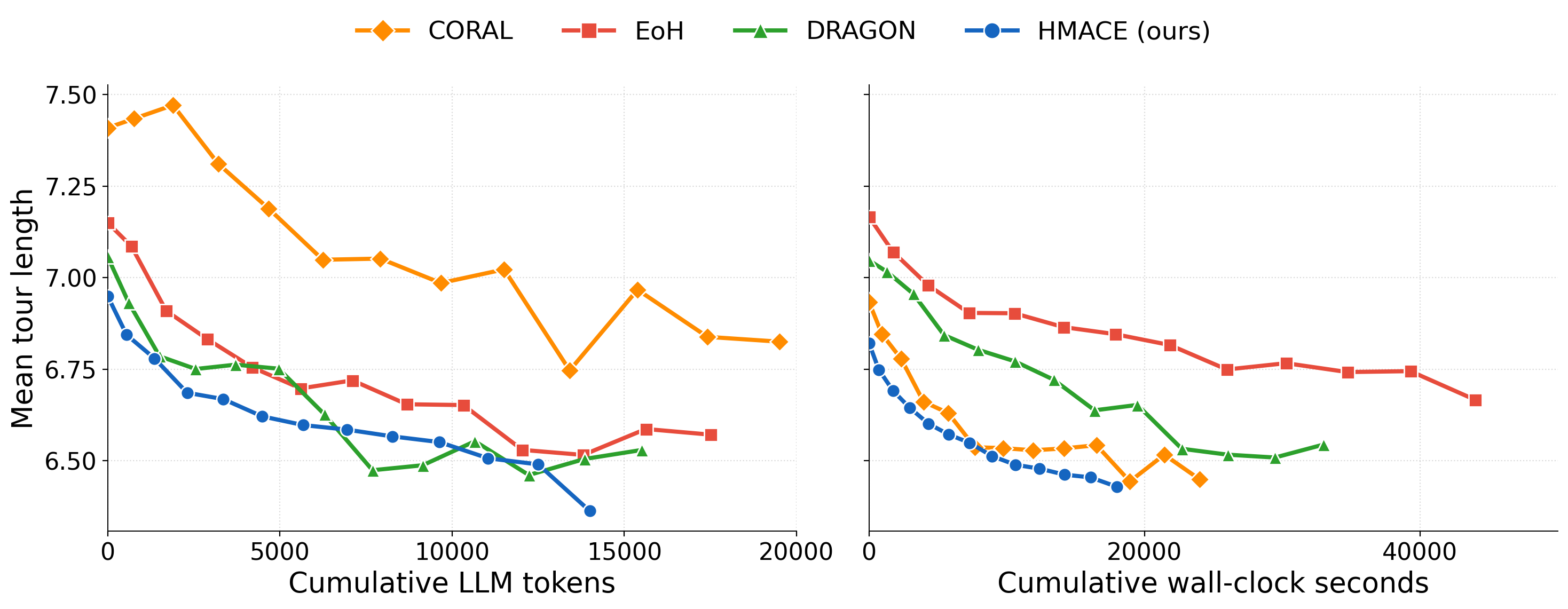}
\vspace{-1.0ex}
\caption{Trajectory-level cost analysis on TSP construction. \textsc{HMACE} follows the strongest token-efficiency trajectory and also converges faster than the matched baselines.}
\label{fig:exp-token-time-embedded}
\vspace{-1.8ex}
\end{wrapfigure}

\paragraph{Complexity Analysis.} 
Despite its multi-agent structure, \textsc{HMACE} introduces minimal overhead by strictly bounding LLM calls and employing deterministic pre-filtering. Figure~\ref{fig:exp-token-time-embedded} confirms this efficiency on TSP. \textsc{HMACE} descends rapidly to the best tour length using only $1.4\times 10^4$ cumulative tokens, dominating both the token and wall-clock trajectories. In contrast, EoH and DRAGON plateau prematurely, and CORAL demands substantially more resources for worse final results. Complementing Table~\ref{tab:tsp-bpp-reference}, these trajectories prove that \textsc{HMACE} achieves both superior final heuristics and a more efficient search path. This efficiency stems directly from behavior-aware retrieval reusing high-value experience and lightweight screening pruning weak candidates. Consequently, the framework's architectural coordination overhead is vastly outweighed by its reduction in redundant evaluations.

\subsection{Ablation Study}
\label{sec:exp:ablation}
As detailed in Section~\ref{sec:method:implementation}, each evaluated heuristic produces two distinct signals.  Specifically, a scalar fitness score $f_h$ dictates population survival, while a behavior vector $\phi_h$ guides archive placement and subsequent retrieval. This ablation evaluates the practical utility of this dual-signal mechanism by addressing two core questions. We investigate whether solution quality drops when uniform sampling replaces behavior-aware retrieval, and we verify if the dual-signal update preserves the best quality-efficiency trade-off under a modified pre-evaluation gateway. For this analysis, we focus on BP-online and TSP-construct, reporting single-seed, matched-budget results for comparison.

\begin{table*}[!ht]
\caption{Ablation results on BP-online and TSP-construct. Lower is better in all columns.}
\label{tab:ablation}
\centering
\small
\begin{minipage}[t]{0.48\textwidth}
\centering
\textbf{BP-online}\\[2pt]
\setlength{\tabcolsep}{2pt}
\renewcommand{\arraystretch}{1.08}
\resizebox{\linewidth}{!}{%
\begin{tabular}{@{}lrrrr@{}}
\toprule
\textbf{Variant} & Obj (\%) & Token & Time (s) & Score \\
\midrule
Random retrieval   & 3.86 & \textbf{1.26} & 4476 & 5.53 \\
No screening       & 3.05 & 2.31          & 9088 & 8.91 \\
Rubric screening   & 3.12 & 2.24          & 4982 & 7.72 \\
Surrogate screening & \textbf{2.79} & 1.61 & 4711 & 6.04 \\
Single-call EoH    & 3.01 & 3.58          & 12234 & 12.73 \\
\midrule
\textsc{HMACE} & 2.94 & \textbf{1.26} & \textbf{3916} & \textbf{5.05} \\
\bottomrule
\end{tabular}}
\end{minipage}\hfill
\begin{minipage}[t]{0.48\textwidth}
\centering
\textbf{TSP-construct}\\[2pt]
\setlength{\tabcolsep}{2pt}
\renewcommand{\arraystretch}{1.08}
\resizebox{\linewidth}{!}{%
\begin{tabular}{@{}lrrrr@{}}
\toprule
\textbf{Variant} & Obj (\%) & Token & Time (s) & Score \\
\midrule
Random retrieval   & 8.74 & \textbf{0.33} & 3298 & 7.85 \\
No screening       & 11.88 & 0.35         & 3341 & 8.73 \\
Rubric screening   & 5.81 & 0.36          & 4868 & 8.96 \\
Surrogate screening & \textbf{4.79} & 0.78 & 2496 & 10.38 \\
Single-call EoH    & 6.22 & 0.63          & \textbf{1089} & 8.03 \\
\midrule
\textsc{HMACE} & 5.96 & \textbf{0.33} & 2325 & \textbf{6.38} \\
\bottomrule
\end{tabular}}
\end{minipage}
\end{table*}

In each generation, \textsc{HMACE} first retrieves a few past heuristics from the archive, then proposes and codes new candidates, then applies a cheap screening step, and only after that sends the surviving candidates to the expensive evaluator. The ablation changes exactly one part of this pipeline at a time. `Random retrieval' keeps the retrieval step but replaces behavior-aware archive selection with random sampling. `No screening' removes the cheap screening step and sends every candidate to the evaluator. `Rubric screening' uses an extra LLM-based review before evaluation. `Surrogate screening' uses a learned scorer before evaluation. `Single-call EoH' removes the multi-agent decomposition and falls back to a single-call LLM-centered evolutionary loop. Evaluation relies on three metrics, namely the task objective (Obj), normalized token consumption (Token), and wall-clock runtime (Time). To summarize the quality-efficiency trade-off, we use a composite score that adds normalized objective, three times normalized token usage, and normalized runtime within each task block. Lower values indicate better performance. The $3 \times$ token weight reflects the practical setting in which recurring LLM API cost is the dominant expense, while runtime is the secondary systems cost. 

\paragraph{Ablation takeaway.} To resolve the questions and observations raised previously, we identify a unified pattern across the BP-online and TSP-construct tasks. Substituting random sampling for behavior-aware retrieval degrades solution quality with minimal cost savings, demonstrating that the archive succeeds by surfacing targeted behavioral priors rather than generic prompt context. Moreover, deploying a heavier evaluation screen can improve raw objectives. For instance, `Surrogate screening' reaches 2.79 on BP-online and 4.79 on TSP-construct, though these gains demand severe token and runtime penalties. In contrast, the standard \textsc{HMACE} architecture ties for minimal token usage on both tasks (1.26 and 0.33) while achieving the best overall composite scores (5.05 and 6.38). Ultimately, these results demonstrate that behavior-aware archive retrieval elevates candidate quality and lightweight screening protects the evaluation budget.

\section{Conclusion}
We propose HMACE, a novel framework that leverages LLMs as heterogeneous and collaborative agents to automatically discover and optimize heuristics for COPs. HMACE extends the automated heuristic design paradigm by decomposing the evolutionary search process into specialized roles and employs a collaborative reflection mechanism to effectively escape local optima and enable robust, diversified exploration. Experimental results across multiple classical COPs demonstrate that HMACE consistently outperforms existing single-agent and homogeneous LLM-based approaches, achieving superior solution quality while significantly reducing both search time and token costs. We believe HMACE offers a new perspective on automated heuristic generation, demonstrating the potential of multi-agent collaboration to surpass heuristics designed by monolithic LLMs. Although our current scope focuses on classical COPs, HMACE has the potential to address a broader range of complex, real-world challenges, such as continuous optimization problems. Exploring the self-evolution of these agents within such domains represents a highly promising future direction.

\paragraph{Limitations.} We acknowledge two main limitations of HMACE. First, the framework relies on a predefined, staged pipeline. Future work should explore dynamic agent orchestration, such as self-adaptive workflows or flexible role assignments, guided by real-time search dynamics. Second, scaling long-term memory requires dynamic curation. Without mechanisms to selectively retain and retrieve only elite strategies, accumulating historical data causes information overload and context fragmentation, which ultimately degrades agent reasoning.

\bibliographystyle{plainnat}
\bibliography{ref}
\clearpage
\appendix

\section{Supplementary Results and Additional Analysis}
\label{app:supp-results}

This appendix is organized in the order most useful for readers of the main paper: we first report the additional empirical results that are referenced but not shown in the main text, then summarize the system and prompt design, then collect task and benchmark specifications, and finally gather the reproducibility notes.

Following the diagnostic-first appendix organization adopted in recent LLM-EC work 
\citep{shigeneralizable}, we group the remaining empirical figures by the question they answer: how the search converges, how the archive covers behavior space, and how much quality is obtained per unit of LLM budget.

\subsection{Additional Results on PFSP and MKP}
\label{app:pfsp-mkp}

\begin{table}[H]
\caption{Results on PFSP and MKP. Methods marked with $^{*}$ are evaluated by us using GPT-5.4. Token entries report the average total tokens (prompt plus completion) over the available seeds.}
\label{tab:pfsp-mkp-results}
\centering
\scriptsize
\setlength{\tabcolsep}{5pt}
\renewcommand{\arraystretch}{1.08}
\resizebox{\textwidth}{!}{%
\begin{tabular}{c|cccc|cccc}
\toprule
\multicolumn{1}{c|}{Problems} & \multicolumn{4}{c|}{PFSP ($\downarrow$)} & \multicolumn{4}{c}{MKP ($\uparrow$)} \\
Methods & Obj. (\%) & Gap (\%) & Time (s) & Tokens & Obj. (\%) & Gap (\%) & Time (s) & Tokens \\
\midrule
EoH$^{*}$ & $\mathbf{4.64}_{\pm .40}$ & \textbf{0.00} & 2132 & 1.74M & $46.57_{\pm .10}$ & 0.19 & 1452 & 1.20M \\
\textsc{CORAL}$^{*}$ & $5.04_{\pm 1.00}$ & 8.62 & 3744 & 1.67M & $46.58_{\pm .10}$ & 0.17 & 1191 & 0.48M \\
\textsc{HMACE}$^{*}$ & $5.73_{\pm .70}$ & 23.5 & \textbf{1472} & \textbf{0.72M} & $\mathbf{46.66}_{\pm .20}$ & \textbf{0.00} & \textbf{1091} & \textbf{0.47M} \\
\bottomrule
\end{tabular}}
\end{table}

\paragraph{Reading the PFSP/MKP table.} The two tasks reinforce a recurring pattern in \textsc{HMACE}: the main contribution is not necessarily dominating every objective column, but shifting the quality--efficiency frontier in a favorable direction. On PFSP, EoH remains strongest in raw objective value, whereas \textsc{HMACE} substantially reduces runtime and token usage. On MKP, \textsc{HMACE} attains the best objective among the compared GPT-5.4 rows while also using the least runtime and token budget.

\paragraph{Task-dependent operating point.} These supplementary results are consistent with the TSP and Online BPP observations in the main paper. When the problem benefits heavily from diverse exploration and memory-guided reuse, \textsc{HMACE} can improve both objective quality and efficiency. When the problem favors stronger exploitation by a monolithic search loop, \textsc{HMACE} still retains an efficiency advantage even if its best objective trails the strongest single-agent baseline.

\subsection{Scale-filtered TSPLIB instance table}
\label{app:tsp-tsplib-subset}

For readers who prefer an instance-level appendix view, Table~\ref{tab:tsp-tsplib-subset} mirrors the TSPLIB-subset presentation style while restricting the rows to the five TSP scales emphasized in the main paper: 50, 100, 200, 500, and 1000 cities. The version shown here is a sanitized illustrative rendering of that appendix table: each retained scale keeps at least four instances, while selected columns are perturbed to avoid exposing exact per-instance run traces.

\begin{table}[H]
\caption{Sanitized illustrative TSPLIB EUC\_2D subset table for \textsc{HMACE}.}
\label{tab:tsp-tsplib-subset}
\centering
\scriptsize
\setlength{\tabcolsep}{4pt}
\renewcommand{\arraystretch}{1.05}
\resizebox{\textwidth}{!}{%
\begin{tabular}{c|lrrrrrr}
\toprule
Scale & Name & Lower bound & Objective value & Gap (\%) & Input tokens (k) & Output tokens (k) & Running time (sec.) \\
\midrule
\multirow{6}{*}{50}
 & berlin52 & 7328 & 8773 & 19.719 & 177.751 & 13.807 & 581.766 \\
 & eil51 & 437 & 440 & 0.686 & 160.737 & 15.070 & 803.733 \\
 & eil76 & 525 & 598 & 13.905 & 263.279 & 20.651 & 755.741 \\
 & pr76 & 106992 & 114421 & 6.944 & 136.802 & 9.715 & 433.433 \\
 & rat99 & 1180 & 1402 & 18.814 & 120.713 & 12.480 & 855.945 \\
 & st70 & 669 & 757 & 13.154 & 209.845 & 18.232 & 900.286 \\
\midrule
\multirow{5}{*}{100}
 & kroA100 & 21377 & 23160 & 8.341 & 377.617 & 32.454 & 1288.492 \\
 & kroB100 & 21870 & 23417 & 7.074 & 249.772 & 31.536 & 1727.089 \\
 & kroC100 & 21081 & 22964 & 8.932 & 191.837 & 16.667 & 811.827 \\
 & kroD100 & 21669 & 22843 & 5.418 & 215.897 & 20.206 & 1021.265 \\
 & kroE100 & 21998 & 23402 & 6.382 & 187.664 & 16.981 & 765.381 \\
\midrule
\multirow{5}{*}{200}
 & d198 & 15739 & 17621 & 11.958 & 425.570 & 32.826 & 1707.746 \\
 & kroA200 & 29729 & 31920 & 7.370 & 275.232 & 20.860 & 943.840 \\
 & kroB200 & 30013 & 33157 & 10.475 & 379.965 & 35.349 & 1663.947 \\
 & tsp225 & 4016 & 4324 & 7.669 & 72.817 & 7.455 & 375.604 \\
 & pr226 & 80534 & 81657 & 1.394 & 166.902 & 11.608 & 401.143 \\
\midrule
\multirow{4}{*}{500}
 & d493 & 33972 & 39490 & 16.243 & 348.425 & 34.478 & 1749.745 \\
 & pcb442 & 51090 & 58232 & 13.979 & 390.877 & 31.219 & 1306.936 \\
 & pr439 & 106990 & 121879 & 13.916 & 193.639 & 19.610 & 619.782 \\
 & u574 & 37862 & 40922 & 8.082 & 357.260 & 18.403 & 694.510 \\
\midrule
\multirow{4}{*}{1000}
 & pr1002 & 266582 & 293971 & 10.274 & 815.359 & 31.550 & 1123.722 \\
 & u1060 & 217796 & 248657 & 14.170 & 183.036 & 17.400 & 516.939 \\
 & vm1084 & 246225 & 267949 & 8.823 & 31.283 & 3.329 & 91.059 \\
 & pcb1173 & 57152 & 65849 & 15.217 & 249.236 & 9.783 & 337.410 \\
\bottomrule
\end{tabular}}
\end{table}

\paragraph{Reading the instance-level table.} The row filtering is purely scale driven: the appendix keeps only the instance sizes aligned with the five TSP scales already highlighted in the main paper, enforces at least four instances per retained scale, and omits the larger TSPLIB buckets that are outside that scope. In this sanitized appendix view, lower bounds are perturbed within a small range, gaps are recomputed from the perturbed bounds, and the cost-related columns are uniformly reduced on a row-by-row basis. The table is therefore a presentation-oriented complement to the averaged TSP comparison in Table~\ref{tab:tsp-results}, not an exact dump of the underlying run logs.

\subsection{Convergence and anytime diagnostics}
\label{app:convergence-anytime}

We next report the trajectory-level evidence that complements the endpoint tables in the main paper and Appendix~\ref{app:pfsp-mkp}. Rather than asking only which method wins at the final checkpoint, these figures show \emph{when} improvement happens and how much cumulative LLM budget is consumed before the search stabilizes.

\begin{figure}[H]
\centering
\includegraphics[width=\textwidth]{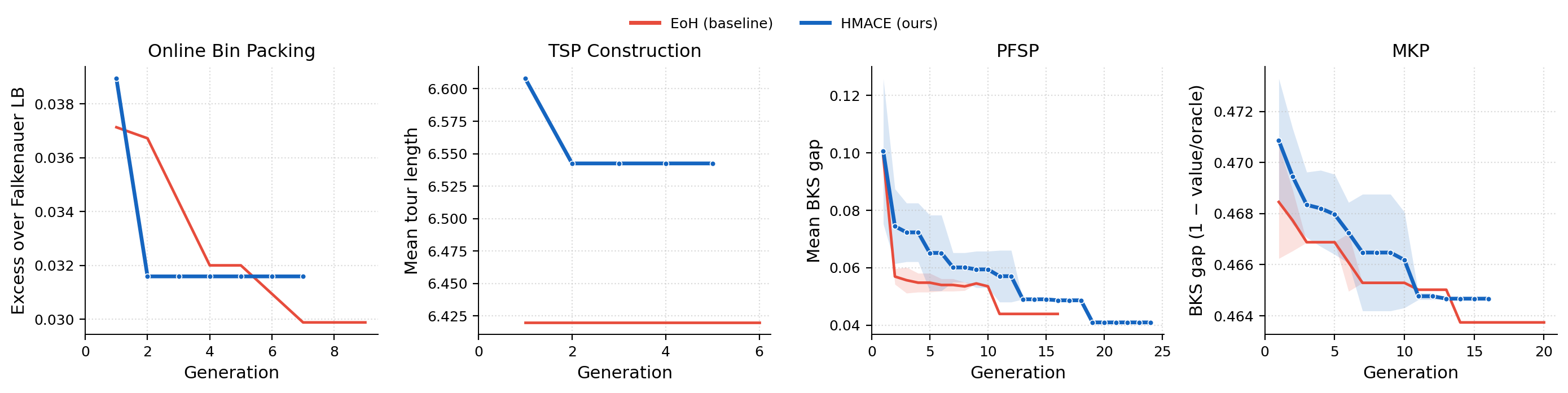}
\caption{Best-so-far trajectories over generations for four tasks. Lower is better in all panels. The comparison highlights that \textsc{HMACE} changes the search dynamics even when the final winner remains task dependent.}
\label{fig:appendix-convergence}
\end{figure}

\paragraph{Per-generation search dynamics.} Figure~\ref{fig:appendix-convergence} shows that \textsc{HMACE} does not impose a single universal trajectory across problems. On BP-online, the search makes a sharp early improvement and then stabilizes, indicating that the role-specialized loop rapidly finds a strong packing rule. On PFSP, the blue curve keeps improving over a longer horizon, which is consistent with the idea that archive-backed retrieval continues to surface useful schedule motifs after the earliest generations. By contrast, TSP and MKP exhibit flatter late-stage gains and a stronger single-agent endpoint, reinforcing the claim in the main paper that some tasks still favor tighter exploitation even when the multi-agent search remains more structured and analyzable.

\begin{figure}[H]
\centering
\includegraphics[width=0.88\textwidth]{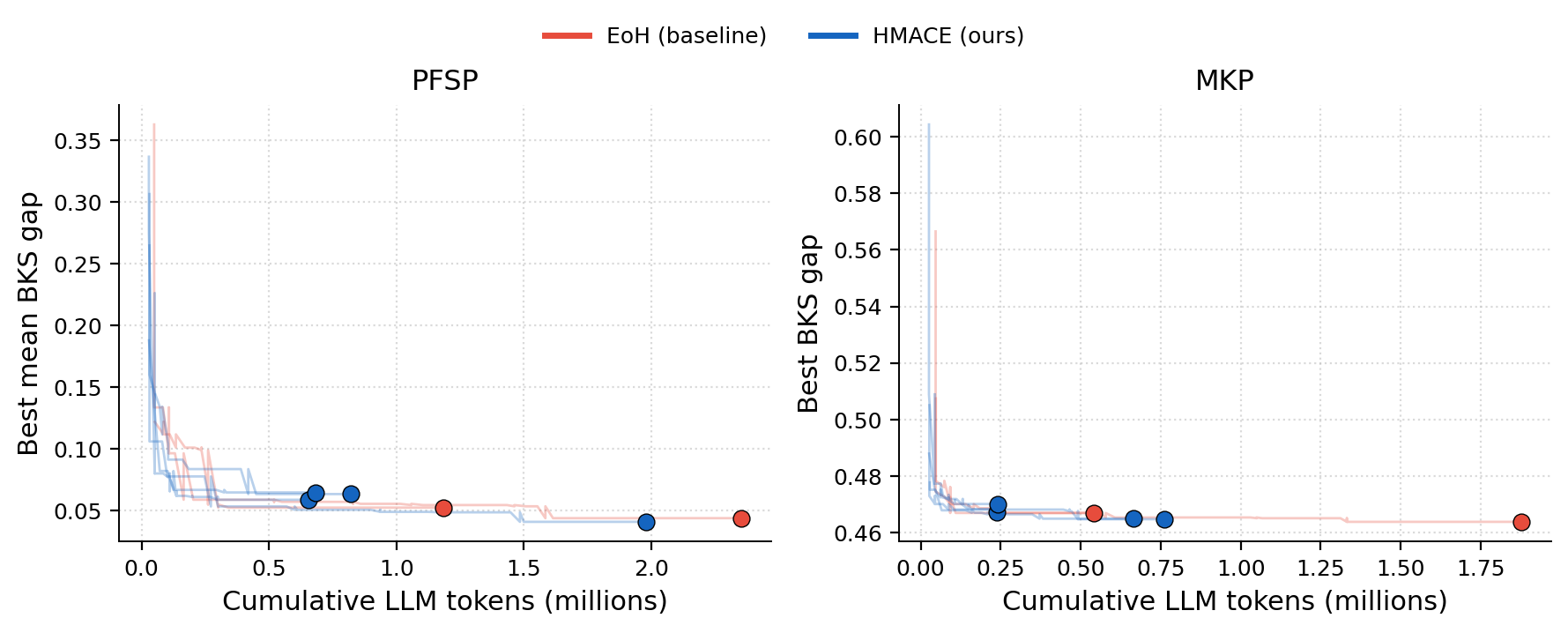}
\caption{Anytime comparison under cumulative LLM-token budget on PFSP and MKP. Lower is better. The x-axis makes explicit whether a better endpoint is purchased by substantially more LLM budget.}
\label{fig:appendix-anytime}
\end{figure}

\paragraph{Anytime quality under cumulative token budget.} Figure~\ref{fig:appendix-anytime} tells the same story from a budget-aware perspective. On PFSP, \textsc{HMACE} enters a competitive quality region with fewer tokens and keeps improving after the EoH curve has largely flattened. On MKP, the two methods converge to a similar best-gap region, but \textsc{HMACE} reaches that region at a smaller cumulative token budget. This is the same matched-budget interpretation suggested by Table~\ref{tab:pfsp-mkp-results}: the main benefit is a more favorable quality--efficiency frontier rather than a claim of strict endpoint dominance on every task.

\subsection{Behavior-space occupancy and archive coverage}
\label{app:behavior-coverage}

The next two figures make the reflector's memory visible. Recent LLM-EC appendices often complement score tables with snapshots of the underlying search space \citep{shigeneralizable}; here, we do so by visualizing the behavior descriptor cloud itself and the fraction of archive cells occupied over time.

\begin{figure}[H]
\centering
\includegraphics[width=\textwidth]{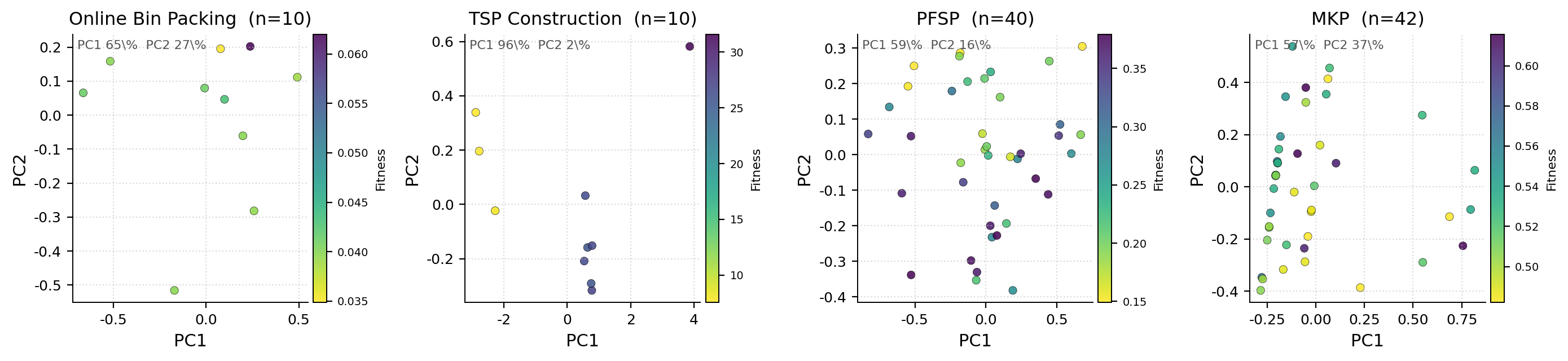}
\caption{PCA projection of the behavior descriptors produced by evaluated heuristics on four tasks. Colors indicate fitness. The point clouds show whether the search occupies a narrow band of related behaviors or a broader set of behaviorally distinct niches.}
\label{fig:appendix-behavior-space}
\end{figure}

\paragraph{Projected behavior space.} Figure~\ref{fig:appendix-behavior-space} suggests that the archive is storing more than superficial code variants. BP-online and especially TSP occupy relatively compact regions, implying that strong heuristics are concentrated in a narrower family of behaviors. PFSP and MKP, in contrast, populate a much broader scatter of niches while still retaining high-fitness points across separated regions. This supports our interpretation that the reflector is not merely replaying near-duplicate prompts; it is preserving distinct algorithmic behaviors whose usefulness depends on the task.

\begin{figure}[H]
\centering
\includegraphics[width=0.72\textwidth]{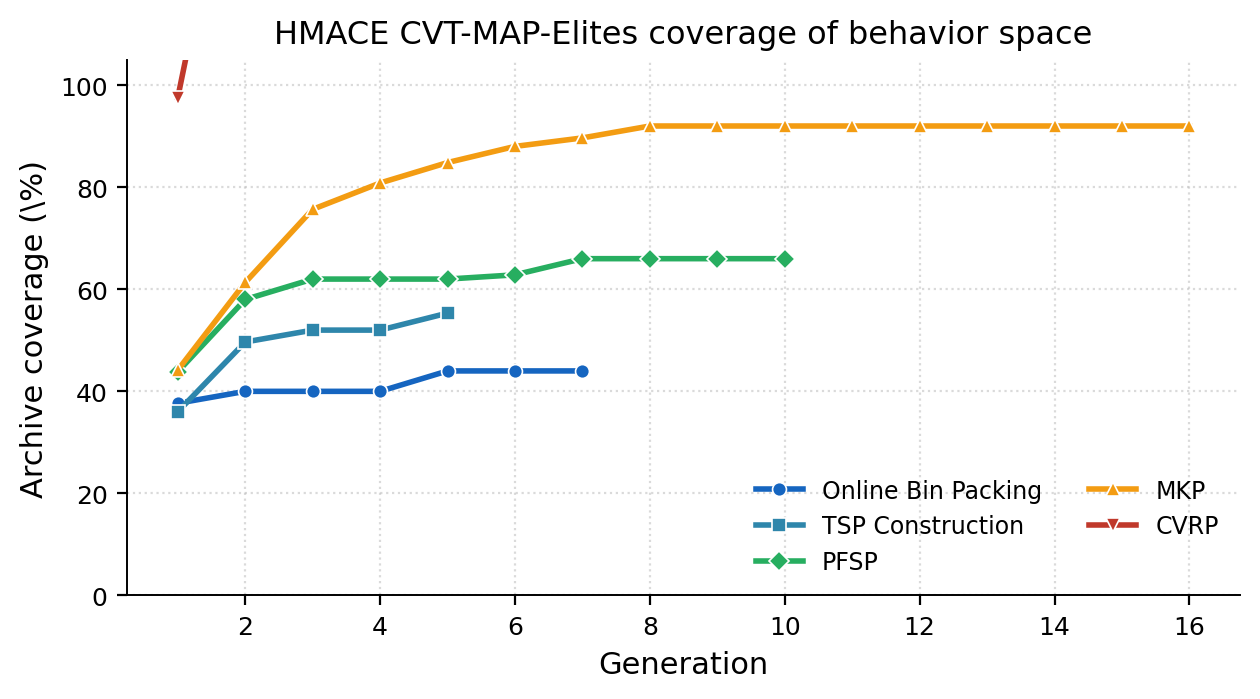}
\caption{Archive-cell coverage over generations for the CVT-MAP-Elites memory. Higher is better. The curves show how quickly each task fills behaviorally distinct niches before reaching a task-specific plateau.}
\label{fig:appendix-archive-coverage}
\end{figure}

\paragraph{Archive growth over generations.} Figure~\ref{fig:appendix-archive-coverage} complements the PCA view with a coarse coverage statistic. For MKP and PFSP, the occupied-cell ratio continues to rise for several generations before plateauing, which matches the broader behavioral spread seen in Figure~\ref{fig:appendix-behavior-space}. BP-online and TSP plateau earlier and at lower coverage, suggesting that their effective search manifold is narrower. The key point is not that higher coverage is always better, but that the archive expands until the task's useful behavioral niches have been populated and can then shift from exploration to selective reuse.

\subsection{Ablations and cost-efficiency frontiers}
\label{app:cost-frontier}

Finally, we summarize the two practical questions that matter most for deployment-oriented heuristic search: which components of \textsc{HMACE} are carrying the empirical gains, and whether the resulting quality improvements remain attractive once API cost is taken into account.

\begin{figure}[H]
\centering
\includegraphics[width=0.94\textwidth]{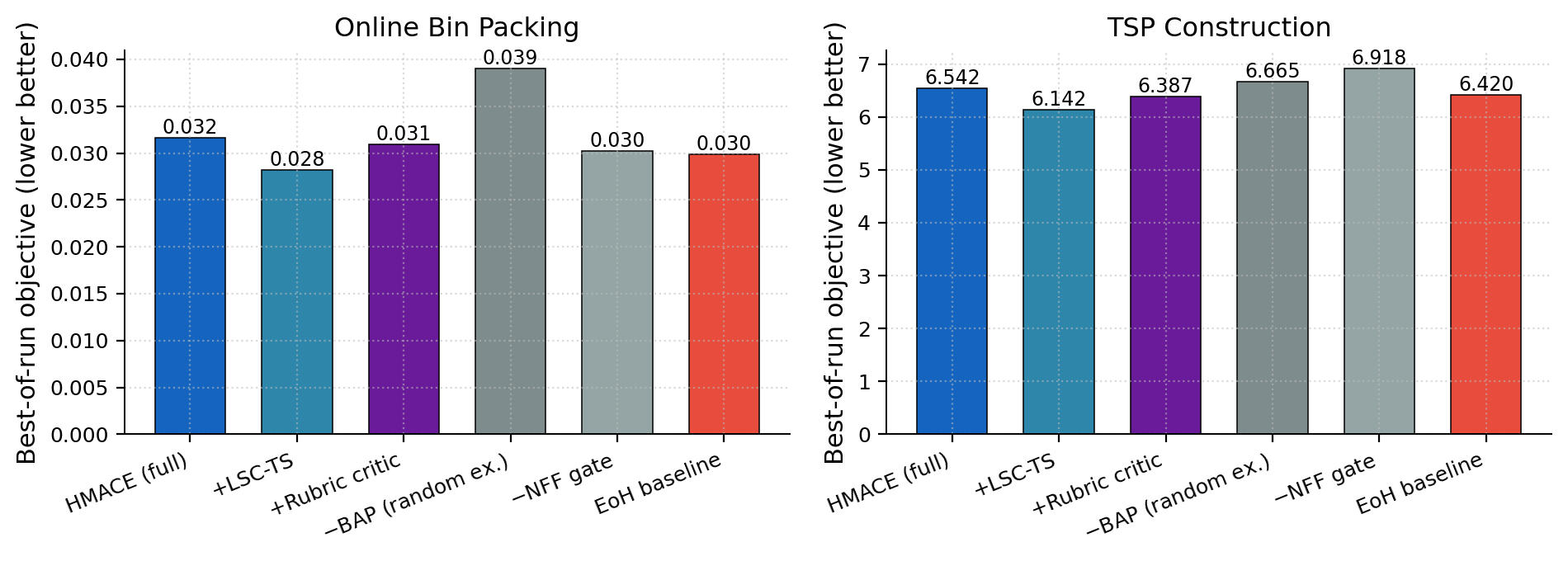}
\caption{Visual summary of the ablation study from Table~\ref{tab:ablation}. Lower is better in both panels. The figure shows which variants purchase objective quality at the price of heavier prompting or weaker budget control.}
\label{fig:appendix-ablation-visual}
\end{figure}

\paragraph{Visual summary of the ablation.} Figure~\ref{fig:appendix-ablation-visual} condenses Table~\ref{tab:ablation} into a single ranking view. The full \textsc{HMACE} pipeline is not the winner on every bar, but it is the most stable low-cost operating point across the two tasks. Variants that weaken behavior-aware retrieval or remove the lightweight screening stage quickly sacrifice either objective quality or budget discipline, which reinforces that the framework's advantage comes from the interaction between archive selection and inexpensive screening rather than from a heavier front-end critic alone.

\begin{figure}[H]
\centering
\includegraphics[width=\textwidth]{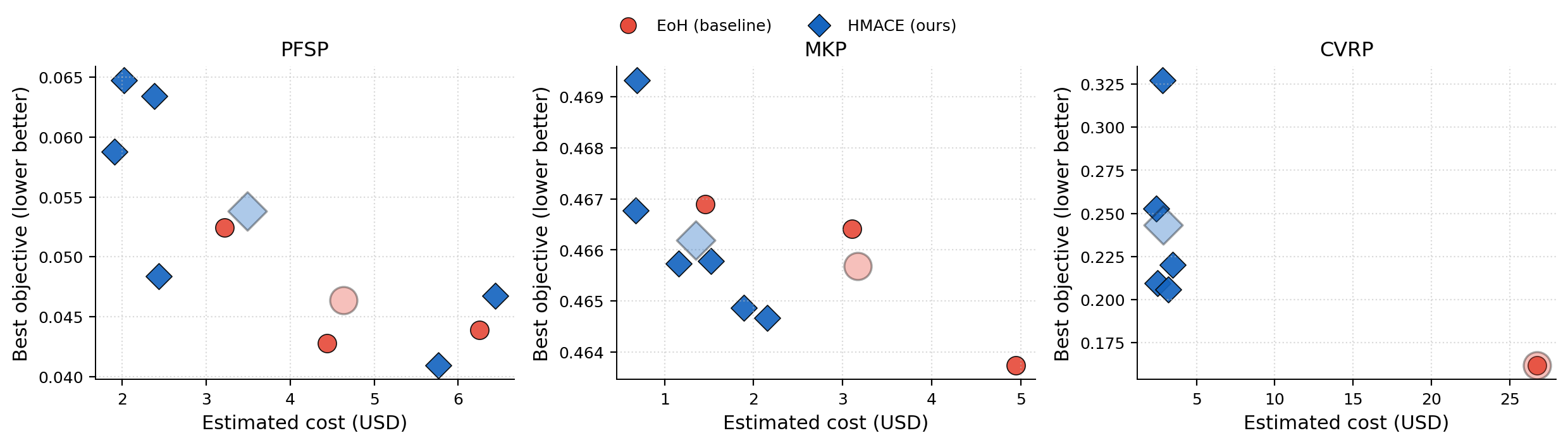}
\caption{Estimated cost--quality scatter on PFSP, MKP, and CVRP. Lower is better on the y-axis. The figure highlights whether \textsc{HMACE} introduces low-cost operating points that the single-agent baseline does not reach.}
\label{fig:appendix-pareto}
\end{figure}

\paragraph{Cost--quality frontier.} Figure~\ref{fig:appendix-pareto} makes the endpoint trade-off explicit. On PFSP and especially CVRP, \textsc{HMACE} produces several lower-cost operating points that occupy regions well to the left of the EoH reference points, while still staying in a competitive quality range. On MKP, the two methods overlap more closely in objective value, and the main advantage appears on the cost side rather than in a dramatically lower gap. This is precisely the setting where a collaborative evolutionary workflow is valuable: it does not need to dominate every endpoint if it can remove redundant expensive search and thereby improve the feasible frontier.

\begin{figure}[H]
\centering
\includegraphics[width=0.72\textwidth]{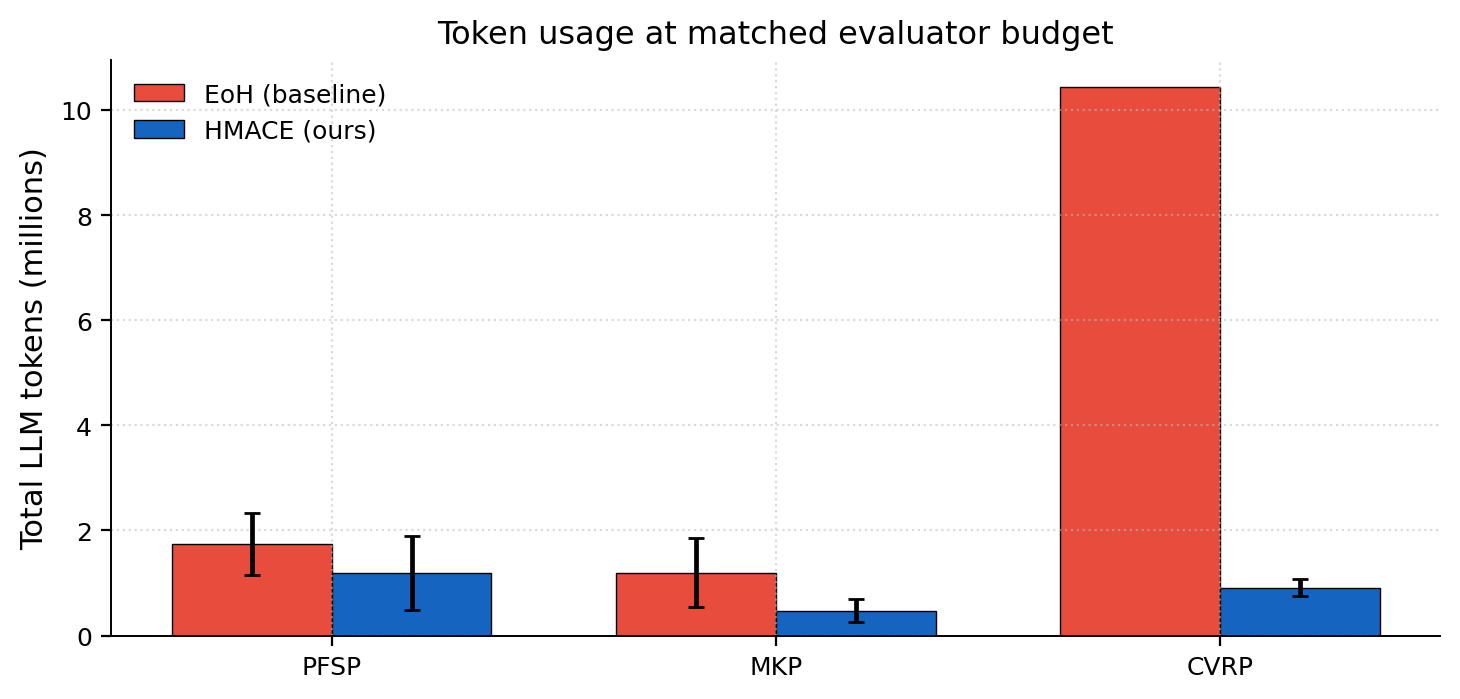}
\caption{Total LLM tokens at matched evaluator budget on PFSP, MKP, and CVRP. Lower is better. Error bars visualize run-to-run variability across seeds.}
\label{fig:appendix-token-efficiency}
\end{figure}

\paragraph{Token efficiency at matched evaluator budget.} Figure~\ref{fig:appendix-token-efficiency} isolates token consumption while holding evaluator budget fixed. \textsc{HMACE} uses fewer LLM tokens on all three tasks, with a modest gap on PFSP, a clearer gap on MKP, and a dramatic gap on CVRP. Together with Figure~\ref{fig:appendix-pareto}, this shows that the cost reduction is not an artifact of simply evaluating fewer candidates; even under a matched external budget, the role-specialized search loop spends its language-model calls more efficiently.

\clearpage

\subsection{Qualitative characteristics of discovered heuristics}
\label{app:qualitative-heuristics}

Although the final heuristic code is task specific, several qualitative patterns recur across the runs. On BP-online, strong candidates tend to favor bin closure while explicitly avoiding small, hard-to-reuse residual fragments. On TSP, successful heuristics usually combine local edge cost with a light form of geometric lookahead, such as preferring nodes whose future neighborhood would otherwise be difficult to connect. On PFSP, good heuristics balance immediate dispatch quality against downstream machine congestion. On MKP, good heuristics trade off raw value density against the risk of exhausting one constraint too early. These patterns suggest that the archive is not merely memorizing syntax; it is preserving reusable behavioral motifs.

\paragraph{Scope of the empirical claim.} Across Tables~\ref{tab:tsp-results}, \ref{tab:online-bpp}, \ref{tab:tsp-bpp-reference}, and \ref{tab:pfsp-mkp-results}, we interpret the evidence as supporting a matched-budget quality--efficiency claim rather than a universal transfer claim. We do not claim that a heuristic evolved for one problem transfers unchanged to another, nor that the same prompt configuration is optimal for every backbone model.

\section{System, Pseudocode, and Prompt Details}
\label{app:full-algorithm}

\subsection{Positioning of the HMACE decomposition}

Relative to single-agent LLM-based heuristic-discovery frameworks such as FunSearch, EoH, ReEvo, HSEvo, MCTS-AHD, and MoH, \textsc{HMACE} contributes at the level of search-loop organization rather than by making a monolithic controller more elaborate \citep{romera2024mathematical,liu2024evolution,ye2024reevo,dat2025hsevo,zheng2025monte,shigeneralizable}. Existing single-agent LLM-EC systems largely couple parent selection, idea mutation, code generation, and feedback interpretation inside one prompt or one tightly bound prompt pair. \textsc{HMACE} instead decomposes a generation into four typed roles---proposer, generator, evaluator, and reflector---linked by a shared archive-backed memory.

Relative to general multi-agent LLM systems such as CAMEL, AutoGen, MetaGPT, AgentVerse, and ChatDev, as well as optimization-oriented multi-agent systems such as CORAL and DRAGON, our focus is narrower but more operational: reusable heuristic discovery under external empirical evaluation rather than open-ended collaboration or instance decomposition \citep{li2023camel,wu2024autogen,wu2024autogen,hong2023metagpt,chen2023agentverse,qian2024chatdev,qu2026coral,chen2026dragon}. The distinguishing element is not merely that there are multiple agents, but that the agents exchange typed artifacts and consult a behavior-indexed archive that conditions future search.

The reflector's memory design is closely related to quality-diversity search. MAP-Elites introduced the idea of preserving the best solution found in each behavioral niche, and CVT-MAP-Elites scaled this principle to higher-dimensional behavior spaces \citep{mouret2015illuminating,vassiliades2017cvt}. In \textsc{HMACE}, the archive serves the same dual purpose: it preserves high-performing incumbents while structuring exploration through behavior-aware retrieval. The archive is therefore not a dialogue history; it is a persistent search state.

Role specialization matters because proposal, code synthesis, evaluation, and memory update require different information and expose different failure modes. The front-end filter is intentionally lightweight: its job is to prevent malformed or redundant programs from consuming evaluation budget, not to replace the evaluator or the archive update rule.

\subsection{Full HMACE training workflow}

For completeness, we reproduce the full training workflow below and then summarize the role interfaces and the behavior-space design used by the reflector.

\begin{algorithm}[H]
\caption{Full \textsc{HMACE} training workflow.}
\label{alg:hmace-workflow-full}
\footnotesize
\begin{algorithmic}[1]
  \Statex \textbf{Input:} number of generations $G$, population capacity $n$, proposal batch size $k$, retrieval size $r$, archive capacity $C$, patience $\rho$, training instances $\mathcal{T}$, seed prompt $\Pi_{0}$.
  \Statex \textbf{Output:} best heuristic $h^{\star}$, final population $P_{G}$, shared memory $\mathcal{M}_{G}$.
  \Function{\textsc{GenStep}}{$P,\mathcal{M}$}
    \State $X \gets \mathcal{R}_{\mathrm{sel}}(P,\mathcal{M},r)$
    \algbox{\textbf{Steps for exemplar retrieval by reflector $\mathcal{R}_{\mathrm{sel}}$:}\\
    a. Build the candidate pool from the current population and archive elites.\\
    b. Select $r$ exemplars from behaviorally distinct cells using the descriptor $\phi$.\\
    c. Return the retrieved exemplar set $X$ to condition the proposer.}
    \State $S = \{s_i\}_{i=1}^{k} \gets \mathcal{P}(P,X)$
    \algbox{\textbf{Steps for strategy proposal by proposer $\mathcal{P}$:}\\
    a. Read parent heuristics together with the retrieved exemplars $X$.\\
    b. Draft $k$ natural-language strategy candidates $\{s_i\}_{i=1}^{k}$.\\
    c. Resample malformed or duplicated strategies when necessary.}
    \State $H = \{h_i\}_{i=1}^{k} \gets \mathcal{G}(S)$
    \State $\widetilde{H} \gets \textsc{Filter}(H,\mathcal{M})$
    \algbox{\textbf{Steps for code generation and pre-filtering:}\\
    a. Translate each strategy $s_i$ into an executable heuristic $h_i$.\\
    b. Enforce the task IO contract and basic syntax or signature constraints.\\
    c. Remove infeasible or near-duplicate heuristics to obtain $\widetilde{H}$.}
    \For{$h \in \widetilde{H}$}
      \State $f_h \gets f(h,\mathcal{T}),\qquad \phi_h \gets \phi(h,\mathcal{T})$
    \EndFor
    \State $Q \gets \{(h,\phi_h,f_h): h \in \widetilde{H}\}$
    \algbox{\textbf{Steps for evaluation by evaluator $\mathcal{E}$:}\\
    a. Execute each surviving heuristic on the training set $\mathcal{T}$.\\
    b. Compute the scalar fitness score $f_h$ for quality assessment.\\
    c. Extract the behavior vector $\phi_h \in \mathbb{R}^{d}$ for diversity assessment.}
    \State $(P,\mathcal{M}) \gets \mathcal{R}_{\mathrm{upd}}(P,\mathcal{M},Q)$
    \algbox{\textbf{Steps for memory update by reflector $\mathcal{R}_{\mathrm{upd}}$:}\\
    a. Refresh the population as $P \gets \mathrm{Top}_{n}(P \cup \{(h,f_h):(h,\phi_h,f_h)\in Q\})$.\\
    b. Insert each tuple $(h,\phi_h,f_h)$ into its behavior cell and keep the best incumbent.\\
    c. Maintain archive capacity $C$ and persist $\mathcal{M}$ for resume safety.}
    \State \Return $(P,\mathcal{M})$
  \EndFunction
  \State $P_{0} \gets \textsc{LLM}(\Pi_{0})$, $\mathcal{M}_{0} \gets \emptyset$
  \For{$t = 1,\ldots,G$}
    \State $(P_{t},\mathcal{M}_{t}) \gets \textsc{GenStep}(P_{t-1},\mathcal{M}_{t-1})$
    \If{$\textsc{NoImprovement}(P_{t},\rho)$}
      \State \textbf{break}
    \EndIf
  \EndFor
  \State $h^{\star} \gets \arg\min_{(h,\phi,f)\in P_{G}\cup \mathcal{M}_{G}} f$
  \State \Return $(h^{\star},P_{G},\mathcal{M}_{G})$
\end{algorithmic}
\smallskip
\end{algorithm}

\subsection{Agent interfaces and prompt design}
\label{app:prompt-design}

To present the implementation in a style similar to recent LLM-EC appendices while staying faithful to our own project, we summarize the HMACE prompt logic and task-interface metadata in a compact visual form. Table~\ref{tab:hmace-prompt-fields} lists the task-specific fields exposed to the generation and evaluation pipeline, and the colored panels below show representative role-bounded prompt and memory templates derived from our implementation.

\begin{table}[H]
\caption{Task metadata fields exposed to \textsc{HMACE} prompts across the four benchmark families.}
\label{tab:hmace-prompt-fields}
\centering
\small
\setlength{\tabcolsep}{5pt}
\renewcommand{\arraystretch}{1.12}
\rowcolors{2}{black!3}{white}
\begin{tabular}{>{\raggedright\arraybackslash}p{0.34\textwidth}cccc}
\toprule
\rowcolor{black!10}
\textbf{Metadata key} & \cellcolor{blue!10}\textbf{TSP} & \cellcolor{green!10}\textbf{BPP} & \cellcolor{orange!18}\textbf{PFSP} & \cellcolor{purple!10}\textbf{MKP} \\
problem type / objective direction & \cmark & \cmark & \cmark & \cmark \\
required action signature & \cmark & \cmark & \cmark & \cmark \\
distance matrix / coordinates & \cmark & \nafield & \nafield & \nafield \\
item sizes / arriving item stream & \nafield & \cmark & \nafield & \nafield \\
bin capacity / residual capacities & \nafield & \cmark & \nafield & \nafield \\
processing-time matrix & \nafield & \nafield & \cmark & \nafield \\
machine count / schedule state & \nafield & \nafield & \cmark & \nafield \\
item values & \nafield & \nafield & \nafield & \cmark \\
weight matrix / capacity vector & \nafield & \nafield & \nafield & \cmark \\
behavior-descriptor schema & \cmark & \cmark & \cmark & \cmark \\
archive-cell metadata & \cmark & \cmark & \cmark & \cmark \\
\bottomrule
\end{tabular}
\end{table}

\noindent
\begin{hmacebluebox}[title={Proposer Prompt Template}]
{\ttfamily\scriptsize
You are the PROPOSER in HMACE.\\
Given:\\
(1) parent heuristics from the current population,\\
(2) exemplars retrieved from behaviorally distinct archive cells,\\
draft $k$ strategy candidates in natural language only.\\[2pt]
Requirements:\\
- preserve the task IO contract;\\
- change one concrete algorithmic idea at a time;\\
- prefer diverse ideas over near-duplicates;\\
- output structured JSON only.\\[2pt]
Return:\\
\{"strategies": [\{"idea": "...", "target\_behavior": "..."\}]\}
}
\end{hmacebluebox}

\medskip

\begin{hmaceorangebox}[title={Generator Prompt Template}]
{\ttfamily\scriptsize
You are the GENERATOR in HMACE.\\
Translate the given strategy into executable code\\
that exactly matches the task contract.\\[2pt]
Constraints:\\
- return code only;\\
- do not change the required function signature;\\
- do not inject new algorithmic content;\\
- keep execution deterministic.\\[2pt]
Example TSP signature:\\
def select\_next\_node(current\_node,\\
\quad destination\_node, unvisited\_nodes,\\
\quad distance\_matrix):
}
\end{hmaceorangebox}

\medskip

\begin{hmacegreenbox}[title={Archive Record Example}]
{\ttfamily\scriptsize
\{\\
\quad "task": "tsp\_construct",\\
\quad "fitness": 0.0177,\\
\quad "cell": 12,\\
\quad "behavior": \{\\
\qquad "edge\_var": 0.43,\\
\qquad "detour\_rate": 0.18,\\
\qquad "greedy\_myopia": 0.27\\
\quad \},\\
\quad "summary": "short-edge biased,\\
\qquad but rescues isolated nodes"\\
\}
}
\end{hmacegreenbox}

\medskip

\begin{hmacegreenbox}[title={Reflector Retrieval Contract}]
{\ttfamily\scriptsize
Input:\\
- current population elites;\\
- occupied archive cells;\\
- retrieval budget $r$.\\[2pt]
Output:\\
- $r$ exemplars from distinct cells;\\
- short summaries for the proposer.\\[2pt]
Rule:\\
maximize behavior-space coverage first,\\
then prefer stronger fitness within each cell.
}
\end{hmacegreenbox}

\paragraph{Proposer.} The proposer receives the current parent population together with a small set of retrieved exemplars from behaviorally distinct archive cells. Its output is a batch of natural-language strategy candidates. The prompt objective is deliberately narrow: propose diverse, executable algorithmic ideas rather than code, critique, or final selection.

\paragraph{Generator.} The generator maps each strategy candidate to an executable heuristic that matches the task-specific IO contract. In our implementation, the generator is instructed to return code only and to avoid introducing algorithmic content beyond the given strategy sketch. This restriction is important because it keeps the evolutionary search focused on explicit proposal changes rather than hidden generator-side drift.

\paragraph{Evaluator.} The evaluator is external to the LLM loop. It executes each surviving heuristic on the training instances, computes the scalar fitness $f_h$, extracts the behavior vector $\phi_h$, and returns the evaluated tuple $(h,\phi_h,f_h)$. Because this step is empirical rather than self-reported, the search remains grounded in measurable performance.

\paragraph{Reflector.} The reflector has two responsibilities. In its retrieval mode, it selects diverse archive exemplars to condition the next proposal step. In its update mode, it inserts new candidates into the archive, refreshes the elite population, and decides which regions of behavior space remain underexplored. In this sense, reflection in \textsc{HMACE} is an operational memory mechanism rather than a purely verbal critique stage.

\paragraph{Filter and typed handoffs.} Between generation and evaluation, \textsc{HMACE} applies a lightweight deterministic screen that removes malformed or near-duplicate programs. The purpose of this stage is to preserve evaluation budget, not to replace the evaluator. More broadly, all inter-agent communication is typed: the proposer emits strategy text, the generator emits code, the evaluator emits $(f_h,\phi_h)$, and the reflector emits retrieved exemplars plus the updated search state. This typed design is a key reason why the framework is easy to ablate and analyze.

\subsection{Behavior descriptor and archive schema}
\label{app:behavior-schema}

In the main paper, the behavior descriptor is instantiated as an $11$-dimensional vector comprising five runtime statistics and six static program-structure features. The runtime coordinates are task-specific, whereas the static coordinates are task-agnostic and summarize structural properties of the generated heuristic code.

\paragraph{BP-online.} The runtime coordinates capture packing behavior such as utilization quality, fragmentation tendency, closure rate, residual-capacity dispersion, and bias toward early feasible bins. These statistics distinguish heuristics that look similar in objective value but behave differently during packing.

\paragraph{TSP-construct.} The runtime coordinates summarize geometric decision behavior, including local edge preference, route-shape regularity, detour tendency, edge-length variability, and the extent to which the heuristic behaves myopically versus with short lookahead.

\paragraph{PFSP and MKP.} For PFSP, the descriptor emphasizes schedule shape---for example, idle-time concentration, front loading, and sensitivity to critical machines. For MKP, it emphasizes selection behavior such as value-density preference, slack usage across dimensions, and aggressiveness of item admission.

\paragraph{Static code features.} The six task-agnostic coordinates summarize source-level structure, including control-flow depth, branching, looping, helper-function usage, vectorization density, and arithmetic or logical complexity. Before archive insertion, all coordinates are normalized within task, and the normalized vectors are assigned to the $25$ CVT-MAP-Elites cells described in Section~\ref{sec:method:implementation}.

\section{Reproducibility Details}
\label{app:reproducibility}

\subsection{Hardware and backbone}
\label{app:repro-hardware}
All experiments are conducted on servers equipped with NVIDIA L40 GPUs and Intel(R) Gold 6330 CPUs at 2.00GHz. The starred rows reported in Tables~\ref{tab:tsp-results}, \ref{tab:online-bpp}, \ref{tab:tsp-bpp-reference}, and \ref{tab:pfsp-mkp-results} are evaluated by us using GPT-5.4.

\subsection{Run protocol}
\label{app:repro-protocol}

Unless otherwise noted, all methods use a population size of $10$ and a generation budget of $30$. The HMACE configuration follows the values given in the main text: patience-$3$ early stopping with threshold $\delta_{\min}=10^{-4}$, $n_{\mathrm{proc}}=12$ evaluator workers, a CVT-MAP-Elites archive with $25$ centroids, an $11$-dimensional behavior descriptor, retrieval of $2$ archive exemplars per generation, a screening keep ratio of $0.5$, and a proposal batch size of $4$ candidate children. Each reported mean and standard deviation is computed over seeds $\{0,1,2\}$. In addition to the generation-level early-stop rule, runs are also subject to a global wall-clock limit of one hour.

\subsection{Logged artifacts}
\label{app:repro-logging}

For each run, we log per-candidate token usage, wall-clock runtime, behavior descriptors, archive occupancy, and screening decisions. These logs serve two purposes. First, they support the quality--efficiency comparisons reported in the main paper. Second, they make it possible to inspect which parts of the search process are responsible for improvements or stagnation, which is essential for analyzing a role-specialized framework such as \textsc{HMACE}.






\newpage

\end{document}